\documentclass[9pt]{elife}

\usepackage{lipsum} 
\usepackage[version=4]{mhchem}
\usepackage{siunitx}
\DeclareSIUnit\Molar{M}
\usepackage{subcaption} 
\usepackage{wrapfig}
\usepackage{booktabs}
\usepackage{multirow}
\usepackage{tabularx}
\usepackage{float}
\usepackage{orcidlink}
\usepackage{stringstrings}
\usepackage{tabulary}
\title{``Garbage In, Garbage Out'' Revisited: What Do Machine Learning Application Papers Report About Human-Labeled Training Data?}

\author[1,\authfn{1}]{R. Stuart Geiger \orcidlink{0000-0001-7215-0532}}
\author[2]{Dominique Cope \orcidlink{0000-0003-2471-772X}} 
\author[2]{Jamie Ip \orcidlink{0000-0001-9952-4987}} 
\author[3,\authfn{1}]{Marsha Lotosh \orcidlink{0000-0003-0362-8947}}
\author[2]{Aayush Shah \orcidlink{0000-0002-7029-2008}}
\author[2]{Jenny Weng \orcidlink{0000-0002-1017-8908}}
\author[1,\authfn{1}]{Rebekah Tang \orcidlink{0000-0003-4563-5192}}
       
\affil[1]{ University of California, San Diego}
\affil[2]{ University of California, Berkeley}
\affil[3]{ Webster Pacific}

\corr{stuart@stuartgeiger.com}{RSG}

\contrib[\authfn{1}]{The majority of the work on this paper was conducted when this author was affiliated with the University of California, Berkeley.}

\contrib{\vspace{.5cm} \textbf{DOI:}  \href{https://doi.org/10.1162/qss\_a\_00144}{doi.org/10.1162/qss\_a\_00144}}

\contrib{\vspace{.5cm} \textbf{Published:} 30 June 2021}

\contrib{\vspace{.5cm} \textbf{Data, code, and protocols:} \href{https://github.com/staeiou/gigo\_qss\_2021}{github.com/staeiou/gigo\_qss\_2021} \href{https://doi.org/10.5281/zenodo.4906636}{doi.org/10.5281/zenodo.4906636}}

\contrib{\vspace{.5cm} \textbf{Peer reviews:} \newline \href{https://publons.com/publon/10.1162/qss\_a\_00144}{publons.com/publon/} \href{https://publons.com/publon/10.1162/qss\_a\_00144}{10.1162/qss\_a\_00144}}

\contrib{\vspace{.5cm} \textbf{Recommended citation:} Geiger, R.S., Cope, D., Ip, J., Lotosh, M., Shah, A., Weng, J., and Tang, R. (2021). ``Garbage In, Garbage Out'' Revisited: What Do Machine Learning Application Papers Report About Human-Labeled Training Data? \textit{Quantitative Science Studies} 2(2). \href{https://doi.org/10.1162/qss\_a\_00144}{doi.org/10.1162/qss\_a\_00144}}

\contrib{\vspace{.5cm} \textbf{Copyright:} (C) 2021 The Authors. Published under the \href{https://creativecommons.org/licenses/by/4.0/}{Creative Commons Attribution 4.0 International license (CC BY 4.0)}}

\begin{document}

\maketitle
\begin{abstract}

Supervised machine learning, in which models are automatically derived from labeled training data, is only as good as the quality of that data. This study builds on prior work that investigated to what extent 'best practices' around labeling training data were followed in applied ML publications within a single domain (social media platforms). In this paper, we expand by studying publications that apply supervised ML in a far broader spectrum of disciplines, focusing on human-labeled data. We report to what extent a random sample of ML application papers across disciplines give specific details about whether best practices were followed, while acknowledging that a greater range of application fields necessarily produces greater diversity of labeling and annotation methods.   Because much of machine learning research and education only focuses on what is done once a "ground truth" or "gold standard" of training data is available, it is especially relevant to discuss issues around the equally-important aspect of whether such data is reliable in the first place. This determination becomes increasingly complex when applied to a variety of specialized fields, as labeling can range from a task requiring little-to-no background knowledge to one that must be performed by someone with career expertise. 
\vspace{.2cm}

\vspace{-.5cm}
\end{abstract}

\vspace{-.25cm}

\section{Introduction}
Supervised machine learning (ML) is now widely used in many fields to produce models and classifiers from training data, which allows for automation of tasks such as: diagnosing medical conditions \citep{ye_predicting_2003,shipp_diffuse_2002}, identifying astronomical phenomena \citep{ball_data_2010,fluke_surveying_2020}, classifying environmental zones \citep{lary_machine_2016,ma_review_2017}, or distinguishing positive versus negative sentiment in documents \citep{prabowo_sentiment_2009,thelwall_sentiment_2010,ravi_survey_2015}. Applying supervised ML requires labeled training data for a set of entities with known properties (called a ``ground truth'' or ``gold standard''), which is used to create a classifier that will make predictions about new entities of the same type. 

``Garbage In, Garbage Out'' is a classic saying in computing about how problematic input data or instructions will produce problematic outputs \citep{mellin1957work,babbage2011passages}, which is especially relevant in ML. Yet data quality is often less of a concern in ML research and education, with these issues often passed over in major textbooks \citep[e.g.][]{friedman2009elements,james2013introduction,goodfellow2016deep}. Instead, the focus is typically on the domain-independent mathematical foundations of ML, with ML education and research often using clean, tidy, and pre-labeled ``toy'' datasets. While this may be useful for theoretically-oriented basic ML research, those applying ML in any given domain must also understand how low-quality or biased training data threatens the validity of the model \citep{buolamwini2018gender,dastin_amazon_2018,obermeyer_dissecting_2019,geiger2020garbage}. 

In this paper, we empirically investigate and discuss a wide range of issues and concerns around the production and use of training data in applied ML. Our team of seven labelers systematically examined published papers that applied supervised machine learning to a particular domain, sampling from three sets of academic fields: life and biomedical sciences; physical and environmental sciences; and social sciences and humanities. For each paper, we asked up to fifteen questions about how the authors reported using supervised ML and how they reported obtaining the labeled training data used to produce the model or classifier. We particularly focus on human-labeled or human-annotated training data, in which one or more individuals make discrete assessments of items. Given that many issues and biases can emerge around human labeling, we examine whether papers reported following best practices in human labeling.

Our project is based on the methodology of structured content analysis, which seeks to systematically turn qualitative phenomenon into categorical and quantitative data \citep{riff2013analyzing}. We draw on and situate our study within the growing efforts to bridge the fields of qualitative and quantitative science studies \citep{leydesdorff_bridging_2020,bowker_numbers_2020,cambrosio_beyond_2020,kang_against_2020}. Quantitative science studies often examines the outputs of science, such as analyzing bibliometrics and other already-quantitative trace data to understand how scientists' final products have been received within science and other institutions. In contrast, qualitative science studies often examines the research process ``in action'' \citep{latour1987science} to investigate case studies how science is produced, such as using more ethnographic or historical methods. This project is in between these two traditions: our method involves systematically quantifying information from qualitative sources, rather than using already-quantitative trace data; we examined a broad set of publications from across domains, rather than more in-depth case studies; and we analyzed and quantified information about research practices, rather than how publications are cited.

As our research project was a human-labeling project studying other human-labeling projects, we took care in our own practices. Before the research project began, we detailed all questions and valid responses, developed instructions with examples, and had a discussion-based process of reconciling disagreements. Another key issue in data labeling are issues of construct validity and operationalization \citep{jacobs_measurement_2019}: is the labeling process actually capturing the theoretical construct that the authors are claiming to capture? In our study, we only have access to the paper reporting about the study and not the actual study or dataset itself. This means our fundamental unit of analysis must be what the papers report, even though our broader intent is to understand what the study's authors and labelers actually did. Many papers either do not discuss such details at all or without sufficient detail to make a determination. For example, many papers did note that the study involved the creation of an original human-labeled dataset, but did not specify who labeled it. For some of our items, one of the most common labels we gave was ``no information.'' This is a concerning issue, given how crucial such information is in understanding the validity of the training dataset, and by extension, the validity of the classifier.

\section{Literature review and motivation}

\subsection{The problem with low-quality and biased training data}

Curating high-quality training datasets for machine learning involves skill, expertise, and care, especially when items are individually labeled by humans. There can be disastrous results if training datasets are taken as a gold standard when they should not be. Supervised ML models are typically evaluated exclusively using a held-out subset of the original training dataset, making systematic flaws in a training dataset difficult to identify or audit within the traditional paradigm of ML. These concerns are particularly pressing when ML is used for deeply subjective and politicized decisions, like in finance, hiring, welfare, and criminal justice. Many ML training datasets have been found to be systematically biased along various axes, including race and gender, which impacts the accuracy of those ML models \citep[e.g.][]{buolamwini2018gender}. In other cases, more subtle issues arise around labels, such as a paper claiming to have produced an ML classifier distinguishing criminals from non-criminals using only facial images, with allegedly overwhelmingly high accuracy \citep{wu_automated_2016}. As Bergstrom and West \citeyearpar{bergstrom_calling_2020} critique, their labels were problematically derived from the source of the photos: criminals were taken from prison mugshots, while non-criminals were taken from professional social network profiles. Because people generally do not smile in mugshots but do smile in profile photos, Bergstrom and West argue that the original team effectively built a smile classifier, but claimed it was a criminality classifier.

In another domain, an exposé \citep{dastin_amazon_2018} reported that Amazon built an internal ML system for hiring that was later scrapped after it was determined to have substantial gender biases. The training dataset used was based on hiring managers' past decisions, where resumes from those hired were given one label and those not hired were given another label. The classifier was thus trained to approximate years of past decisions, and given that Amazon has had significant gender gaps in their workforce (like many tech companies), this meant such systematic biases were reinforced and rationalized through ML. This is the case even though gender was intentionally excluded as a feature in the model, as the classifier used other features that were a proxy for gender to more closely approximate the biases in the training data. Had the training data been a new dataset labeled by a diverse team of trained HR professionals tasked with evaluating resumes with a focus on non-discrimination, this may have produced a quite different classifier.

Machine learning in the field of medicine is poised for explosive growth, although critics raise similar concerns about training data. Medical privacy risks arise for patients whose health care records may be used in formulating a training dataset \citep{vayena_machine_2018}. Furthermore, there is evidence of biases in health care applications of ML, and in some instances, the consequences of biases may directly impact patients' survival. One study in the U.S. labeled patients medical records with their severity of illness, using a proxy variable that ostensibly required little human judgment: the cost of the patient's healthcare. Yet when this data was used to train a classifier, it caused significant bias against African American patients, who historically have had differential access to medical care \citep{obermeyer_dissecting_2019}. The medical field itself is encountering new questions surrounding human labeling and annotation. For example, one widely used application is the interpretation of medical imaging. The human who labels MRI images as cancerous or not-cancerous must have specific expertise, versus someone who labels product reviews as positive or negative. Meta-research in radiology has found practicing radiologists have about a 3-5\% error rate \citep{brady_error_2016}, which raises the question about whether radiology training datasets should be independently labeled by multiple experts to ensure data quality. Finally, as with many fields, the introduction of ML using pre-existing data from a particular environment and setting has the distinct potential to reproduce and perpetuate existing systemic biases, especially when that classifer is deployed to a different environment and setting \citep{decamp_latent_2020}.

\subsection{``Garbage in, garbage out'' version 1}

This project is heavily based on a prior study \citep{geiger2020garbage}, which similarly had a team of labelers examine issues around training data in a random sample of published papers. That study examined a narrow subset of peer-reviewed and preprint papers in a specific field: applied ML papers trained on Twitter data. They looked for 13 pieces of information in each paper, which they argued were important to understanding the validity of the training data labeling process. This included if the data was human or machine labeled, who the labelers were, how many labelers rated each item, and rates of inter-rater reliability (if multiple labelers rated each item). The study found a wide divergence both in the level of information reported and in adherence to best practices in human labeling. For example, of papers reporting a new human-labeled training dataset, about 75\% gave some information about who the labelers were, 55\% specified the number of labelers, 11\% released the training dataset itself, and 0\% reported how much crowdworkers were paid for their work.

We expanded on Geiger et al's study, drawing heavily from their published questions and protocols. We followed the same general process of having labelers rate each item independently, then reconciled disagreements through a discussion led by the team leader. We made some small modifications and extensions to the questions, which were recommended by the original authors for future work or were better suited to the expanded scope. We added questions about the field/domain of the paper and about the reconciliation process when multiple labelers labeled each item. We also rewrote some of the labeling instructions, label categories, and provided examples, often to clarify ambiguities. 

\subsection{Best practices in human labeling of training data}
Geiger et al \citeyearpar{geiger2020garbage} gives a substantial review of existing work around human labeling of training data, including an extensive discussion of best practices in this work. They argue that much of labeling work for ML is a form of structured content analysis, which is a methodology long used in the humanities and social sciences to turn qualitative or unstructured data into categorical or quantitative data. This involves teams of ``coders'' (also called ``annotators'', ``labelers'', or ``reviewers'') who ``code'', ``annotate'', or ``label'' items individually. (Note that we use label/labeler in this paper, although we began with using annotate/annotator, which is still present in some of our data and protocols.) One textbook describes content analysis as a ``systematic and replicable'' \citep[p. 19]{riff2013analyzing} method with established best practices, as Geiger et al summarizes:

\begin{quote}
A ``coding scheme'' is defined, which is a set of labels, annotations, or codes that items in the corpus may have. Schemes include formal definitions or procedures, and often include examples, particularly for borderline cases. Next, coders are trained with the coding scheme, which typically involves interactive feedback. Training sometimes results in changes to the coding scheme, in which the first round becomes a pilot test. Then, labelers independently review at least a portion of the same items throughout the entire process, with a calculation of ``inter-rater reliability'' (IRR) or ``inter-annotator agreement'' (IAA). Finally, there is a process of ``reconciliation'' for disagreements, which is sometimes by majority vote without discussion and other times discussion-based. \citep[][p. 2-3]{geiger2020garbage}
\end{quote}

Structured content analysis is a difficult task, requiring both domain-specific expertise about the phenomenon to be labeled and domain-independent expertise to manage teams of labelers. Historically, undergraduate students have often performed such work for academic researchers. With the rise of crowdwork platforms like Amazon Mechanical Turk, crowdworkers are often used for data labeling tasks. New software platforms have been developed to support more micro-level labeling and annotation or labeling at scale, including in citizen science \citep{chang_revolt_2017,perez_marky_2015,bontcheva_gate_2013,doccano}. For example, the Zooniverse \citep{Simpson2014} provides a common platform for citizen science projects across domains, where volunteers label data under scientists' direction.

\subsection{Meta-research and methods papers in linguistics and NLP}
\label{sec:mrmethods}
We also draw inspiration from meta-research and standardization efforts in Linguistics and Natural Language Processing (NLP) \citep{bender2018data,McDonald2019}. These fields have developed extensive literatures on standardization and reliability of linguistic labels, including best practices for corpus annotation \citep[e.g.][]{hovy2010towards,doddington2004automatic,linguistic2008ace}. In Geiger et al's 2020 study, the publisher with the highest information score was the Association for Computational Linguistics. There has been much work in linguistics and NLP around these issues, such as Sap et al's study of racial bias among labelers \citep{sap_risk_2019}. Blodgett et al conducted a content analysis of how 146 NLP researchers discuss ``bias'' and found that while this has become a prominent topic in NLP, papers' discussions of motivations and methods around bias ``are often vague, inconsistent, and lacking in normative reasoning'' \citep[][p. 5454]{blodgett2020language}. There is also related work in methods papers focused on identifying or preventing ``low-effort'' responses from crowdworkers \citep{Mozetic2016,soberon2013measuring,raykar2012eliminating}, which raise issues around fair labor practices and compensation \citep{silberman2018responsible}.   

\subsection{The open science, reproducibility, and research integrity movements}
\label{sec:datadoc}
Two related movements in computationally-supported knowledge production have surfaced issues around documentation. First, open science is focused on broader availability to the products of research and research infrastructure, including open access to publications, software tools, datasets, and analysis code \citep{fecher_open_2014}. The related reproducibility movement calls for researchers to make protocols, datasets, and analysis code public, often focusing on what others need to replicate the original study  \citep{Wilson2017, Kitzes2018}. Such requirements have long been voluntary, with few incentives to be a first mover, but funding agencies and publications are increasingly establishing such requirements \citep{Goodman2014,gil_toward_2016}. 
 
One notable effort is around formally specifying what each author of a paper actually did, which has long been standard in medical journals \citep{rennie_contributions_2000}. Author role documentation has gained popularity with the more recent Contributor Roles Taxonomy Project (or CRediT) \citep{brand_beyond_2015}. CRediT declarations are increasingly required by journals, which has led to novel quantitative science studies research \citep{lariviere_investigating_2020}.  We also draw inspiration from work about capturing information in ML data flows and supply chains \citep{singh_decision_2019,schelter_automatically_2017,gharibi_automated_2019} and developing tools to support data cleaning \citep{schelter_automating_2018,krishnan_activeclean_2016}. We note that this work has long been part of library and information science, particularly in Research Data Management \citep{schreier2006academic, borgman2012conundrum, Medeiros2017, sallans2012dmp}. There is much more work to be done on quantitatively studying issues around research integrity \citep{zuckerman_is_2020,silberman2018responsible}, which institutionally has often been limited to more egregious and blatant cases of plagiarism and fabrication. 

\subsection{Fairness, Accountability, and Transparency in Machine Learning}

Within the field of machine learning, there is a growing movement in the Fairness, Accountability, and Transparency (or FAccT) sub-field, with many recent papers proposing training data documentation in the context of ML. Various approaches and metaphors have been taken in this area, including ``datasheets for datasets'' \citep{gebru2018datasheets}, ''model cards'' \citep{mitchell2019model}, ``data statements'' \citep{bender2018data}, ``nutrition labels'' \citep{holland2018dataset}, a ``bill of materials'' \citep{barclay2019towards}, ``data labels'' \citep{beretta2018ethical}, and ``supplier declarations of conformity'' \citep{hind2018increasing}. Many go far beyond the concerns we have raised around human-labeled training data, as some are also (or primarily) concerned with documenting other forms of training data, model performance and accuracy, bias, considerations of ethics and potential impacts, and more. Our work is strongly aligned with this movement, as we seek to include data labeling within these areas of concern. However, as we discuss in our conclusion, a single one-size-fits-all standard may be necessary but not sufficient to address concerns of fairness and bias.

We also call attention to those developing methods for ``de-biasing'' machine learning, which is a fast-moving and contentious research area (for surveys and comparative work, see \citep{mehrabi_survey_2019,friedler_comparative_2019}). Much of this work is in developing domain-independent fairness metrics for evaluating trained models \citep[e.g.][]{hardt_equality_2016,zafar_fairness_2017}, which are used to modify trained models or predictions \citep[e.g.][]{amini_uncovering_2019,karimi2020}. However, other work has approached these issues more as a problem of dataset pre-processing \citep{calmon_optimized_2017} or database repair \citep{salimi_database_2020}. Critics note that domain-independent approaches may fall into what Selbst et al identify as ``abstraction traps'' \citep[][p.60]{selbst_fairness_2019}, such as failing to account for the particularities of different kinds and qualities of discrimination in a given social context --- a critique Hanna et al \citeyearpar{hanna_towards_2020} make of fairness research that treats race as a single fixed attribute. We did not ask any questions about how papers discuss de-biasing or data cleaning due to the large number of questions we were already asking and the novelty of such approaches, but these concerns are deeply related. 

\section{Data and methods}
\subsection{Data: machine learning papers performing classification tasks}
Our goal was to find a corpus of papers using supervised ML across disciplines and application domains, including papers producing an original labeled dataset using human labeling. We used the Scopus bibliographic database \citep{baas_scopus_2020}, which contains about 40,000 publications that a review board has verified for various qualities, including being peer reviewed, regularly published for at least 2 years, and governed by a named editorial board of experts. We searched for journal articles and conference proceedings from 2013 to 2018 where the title, abstract, or keywords included ``machine learning'' and either ``classif*'' or ``supervi*'' (case insensitive). We ran three stratified samples across Scopus's Subject Area classifcations\footnote{\url{https://web.archive.org/web/20200812203800/https://service.elsevier.com/app/answers/detail/a_id/15181/supporthub/scopus/}}: Physical Sciences (which includes engineering and earth/ecological sciences); Social Sciences \& Humanities (a single category); and Life Sciences \& Health Sciences (two categories, which we combined). Table \ref{table-corpus} describes our sampling. More details about the corpora are in the appendix, which is available as supplementary materials and in our data repository (see section \ref{sec:materials}):

\begin{table}[h]
\begin{tabular}{l|l|l|l}
\hline
Corpus & Papers in corpus & \# randomly sampled & \% sampled \\ \hline
Social Sciences \& Humanities & 5,346 & 70 & 1.30 \% \\
Life \& Biomedical Sciences & 9,507 & 60 & 0.63 \% \\
Physical \& Environmental Sciences & 11,030 & 70 & 0.63 \% \\ \hline
Total & 25,883 & 200 & 0.77 \% \\ \hline
\end{tabular}
\caption{Summary of sampling across all three corpora}
\label{table-corpus}
\end{table}
\vspace{-.8cm}
\subsection{Labeling team, training, and workflow}

Our labeling team included one research scientist who led the project (RSG) and undergraduate research assistants, who worked 6-10 hours per week for course credit as part of a university-sponsored research experience program (DC, JI, ML, AS, JW, and RT). The project began with six students for one semester, five of whom continued on the project for the second semester. All students had some coursework in computer science and/or data science, with a range of prior experience in machine learning in both a classroom and applied setting. Students' majors and minors included Electrical Engineering \& Computer Science, Data Science, Statistics, Economics, Linguistics, and Biology. For the first four weeks, the team leader trained the students in both broader ML concepts and the specific questions to be answered for this project. The team first labeled and discussed a practice set of 40 papers sampled from across the three corpora, which were not included in the final dataset. In these initial weeks, the team learned the coding schema and the reconciliation process, which were further refined. 

Following this training, the labeling workflow was that each week, a set of papers were randomly sampled from one corpus, typically between 10-15 papers. The students independently reviewed and labeled the same papers, using different web-based spreadsheets to record labels. The team leader synthesized labels and identified disagreement. The team met in person or by videochat to discuss the week's cases of disagreement. The team leader explained various issues in question and built a consensus about the proper label (as opposed to purely majority vote). The team leader had the final say when a consensus could not be reached. 

All 200 papers were labeled by at least four labelers; one labeled 137 items and another labeled 100 items. Following the first round of labeling and reconciliation, we conducted a second round of verification. Where there was any initial disagreement on labels in the first round, each paper was re-examined and discussed by at least two labelers and the team leader. The second round began multiple months after the first round, meaning that there was at least one month between when each paper was examined and re-examined. If there was still disagreement, the final decision was made by the team leader. The team leader did a final check to review every label for all 200 papers. 

\subsection{Inter-rater reliability and labeled data quality}

All human labeling projects that involve multiple labelers should evaluate the intersubjective reliability of the labeling process \citep{tinsley1975interrater}. We present inter-rater reliability (IRR) metrics using three metrics. For all metrics, we re-coded ``unsure'' and blank responses to both be blank (NaN), but treated ``N/A'' answers as a distinct judgment. First, we calculated mean total agreement, or the proportion of items where all labelers initially gave the same label before reconciliation, but not counting blank \& unsure responses. As Table \ref{irr-example} shows, this is a more stringent metric: all non-blank/unsure responses must be the same for an item to have a 1 score, otherwise the score is 0. Second, we present the mean percent correct rate, which is the proportion of labelers who initially gave the same label arrived after discussion \& reconciliation, but also not counting blank \& unsure responses. As Table \ref{irr-example} shows, this is a more forgiving metric: if 5 out of 6 labelers give the same final correct label, the score is 0.83 for that item. For these two metrics, we calculated per-question scores by taking the mean of all scores for an item. 

\begin{table}[h]
\caption{Example of IRR calculations for sample rows. }\label{irr-example}
\begin{tabular}{|p{.4cm}|p{1cm}|p{1cm}|p{1cm}|p{1cm}|p{1cm}|p{1cm}|p{1.2cm}|p{1cm}|p{1.25cm}|}
\hline
\textbf{\#} & \textbf{Labeler \#1} & \textbf{Labeler \#2} & \textbf{Labeler \#3} & \textbf{Labeler \#4} & \textbf{Labeler \#5} & \textbf{Labeler \#6} & \textbf{Final / correct label} & \textbf{Total agreement} & \textbf{Mean percent correct} \\ \hline
\textbf{1} & yes & unsure & yes & yes & yes & {[}blank{]} & yes & 1 & 1 \\ \hline
\textbf{2} & yes & yes & yes & yes & no & yes & yes & 0 & 0.83 \\ \hline
\textbf{3} & no & n/a & yes & no & yes & {[}blank{]} & yes & 0 & 0.4 \\ \hline
\textbf{4} & yes & no & yes & {[}blank{]} & yes & unsure & no & 0 & 0.25 \\ \hline
\end{tabular}
\end{table}

We also present the widely-used Krippendorff's alpha \citep{krippendorff1970} metric, although we strongly advise against relying on it. Our data does not meet the statistical assumptions for both Fleiss's kappa and Krippendorf's alpha, which are popular because they support missing labels for 3+ labelers and take into account the possibilities that raters made decisions based on random chance. However, this requires assuming a uniform prior possibility of such a random distribution, which generally only applies if each possible response by raters is equally likely. Rates can be dramatically lower when there is a highly skewed distribution of response categories \citep{quarfoot_how_2016,oleinik_choice_2014}. Our dataset has highly skewed distributions, especially for many of the more specialized questions, which lead to miniscule scores for some questions with especially skewed distributions (e.g. prescreening for crowdwork; reported inter-rater reliability).

\begin{table}
\caption{Inter-rater reliability metrics per question. }\label{irr}
\begin{tabular}{|p{5.8cm}|p{2cm}|p{2.5cm}|p{2.1cm}|}
\hline
\textbf{Question} & \textbf{Mean total agreement} & \textbf{Mean percent correct} & \textbf{Krippendorff's alpha} \\ \hline
\textbf{Original classification task} & 66.0\% & 84.8\% & 0.670 \\ \hline
\textbf{Classifier area/domain} & 34.5\% & 65.4\% & 0.520 \\ \hline
\textbf{Labels from human judgment} & 37.5\% & 68.2\% & 0.517 \\ \hline
\textbf{Human labeling for training data} & 46.5\% & 77.3\% & 0.517 \\ \hline
\textbf{Used original human labeling} & 43.5\% & 71.0\% & 0.498 \\ \hline
\textbf{Original human labeling source} & 43.5\% & 71.1\% & 0.330 \\ \hline
\textbf{Prescreening for crowdwork} & 58.5\% & 84.2\% & 0.097 \\ \hline
\textbf{Labeler compensation} & 46.0\% & 68.0\% & 0.343 \\ \hline
\textbf{Training for human labelers} & 48.0\% & 70.0\% & 0.364 \\ \hline
\textbf{Formal instructions} & 47.5\% & 66.8\% & 0.337 \\ \hline
\textbf{Multiple labeler overlap} & 48.5\% & 69.3\% & 0.370 \\ \hline
\textbf{Synthesis of labeler overlap} & 53.0\% & 83.4\% & 0.146 \\ \hline
\textbf{Reported inter-rater reliability} & 55.5\% & 85.8\% & 0.121 \\ \hline
\textbf{Total num of human labelers} & 50.5\% & 69.3\% & 0.281 \\ \hline
\textbf{Median num of labelers per item} & 48.5\% & 69.3\% & 0.261 \\ \hline
\textbf{Link to dataset available} & 41.0\% & 66.1\% & 0.322 \\ \hline
\textbf{Average across all questions} & 48.0\% & 73.1\% & 0.356 \\ \hline
\textbf{Median across all questions} & 48.0\% & 70.0\% & 0.343 \\ \hline
\end{tabular}
\end{table}

Table \ref{irr} presents both our custom metrics and Krippendorff's alpha for all questions. Mean total agreement rates ranged from 37.5\% to 66\%, with an average of 48.0\% across all questions. Mean percent correct rates ranged from 65.4\% to 85.8\%, with an average of 73.1\% across all questions. Some questions that had lower rates (especially for mean total agreement) were due to a labeler making an incorrect assessment on an earlier question, which determines whether they answer subsequent questions or mark them as 'N/A'. 

In interpreting these metrics, we note that the standard approach of human labeling checked by IRR metrics treats individual humans as scientific instruments that turn complex phenomena into discrete structured data. If there is a high degree of inter-rater reliability, then reconciliation can easily take place through a majority vote process involving no discussion, or if rates are quite high, then many researchers assume they can use just one of those human labelers per item in future work. These rates were not high enough for us to have confidence that we could have a purely quantitative / majority-vote reconciliation process, much less a process of only using one labeler per item. However, these rates are sufficient to show there is enough agreement to proceed to a discussion-based reconciliation process and a final check of all items by the team leader. As McDonald et al \citeyearpar{McDonald2019} discuss, standardized IRR metrics like Krippendorf's alpha are useful in highly-structured labeling projects that do not have a discussion-based reconciliation process, as they only evaluate the agreement of independent initial labels. Such metrics would be more essential to the validity of our study if we were conducting a quantitative, majority-rule reconciliation process or if only a subset of items were reviewed by multiple labelers. We included mean percent correct rates to partially include the reconciliation and verification process.

Furthermore, our approach was largely focused on identifying the presence or absence of various kinds of information within long-form publications. This is a different kind of human judgment than is typically involved in common tasks using human labeling for ML (e.g. labeling a single social media post for positive/negative sentiment) or traditional social science and humanities content analysis (e.g. categorizing newspaper articles by topic). Our items were full research publications with many pages of detail, which followed many different field-specific conventions and genres. Our labelers were looking for up to 15 different kinds of information per paper, each of which could be found anywhere in the paper. We reflected that in our reconciliation process, most of the time when labelers disagreed, it was because some had caught a piece of information in the paper that others had not seen. Once that information was brought to the group, it was most often the case that some labelers said that they had missed that information and changed their response. It was less common for our team to have disagreements arising from two labelers differently interpreting the same text, especially after the first few weeks. For such reasons, we are relatively confident that if, after our process, no individual member of our team has identified the presence of such information, then it is quite likely it is not present in the paper.

\subsection{Software, datasets, and research materials}
\label{sec:materials}
We used Google Sheets to enter labels. For computational analysis and scripting for corpus collection, data management, and data analysis, we used Python 3.7 \citep{python}, using the following libraries: Pandas dataframes \citep{pandas} for data parsing and transformation; SciPy \citep{scipy} and NumPy \citep{numpy} for quantitative computations; Matplotlib \citep{Matplotlib} and Seaborn \citep{seaborn} for visualization; and simpledorff \citep{simpledorff} for IRR calculations. Analysis was conducted in Jupyter Notebooks \citep{jupyter} using the IPython \citep{ipython} kernel. 

Datasets, analysis scripts, labeling instructions, and other supplementary information can be downloaded from GitHub\footnote{ \url{https://github.com/staeiou/gigo_qss_2021}} and Zenodo.\footnote{ \url{https://doi.org/10.5281/zenodo.4906636}} Datasets include all labels from all labelers for the first round of independent labeling and the consolidated set of final labels and scores for all items. Paper URLs/DOIs have been anonymized with a unique salted hash. Analysis scripts are in Jupyter Notebooks and can be explored and modified in any modern web browser using the cloud-based MyBinder.org \citep{binder2018}.\footnote{ \url{https://mybinder.org/v2/gh/staeiou/gigo_qss_2021/HEAD}}

\section{Findings}
\begin{figure}[b!]
\begin{fullwidth}

    \centering
    \includegraphics[width=1.3\textwidth]{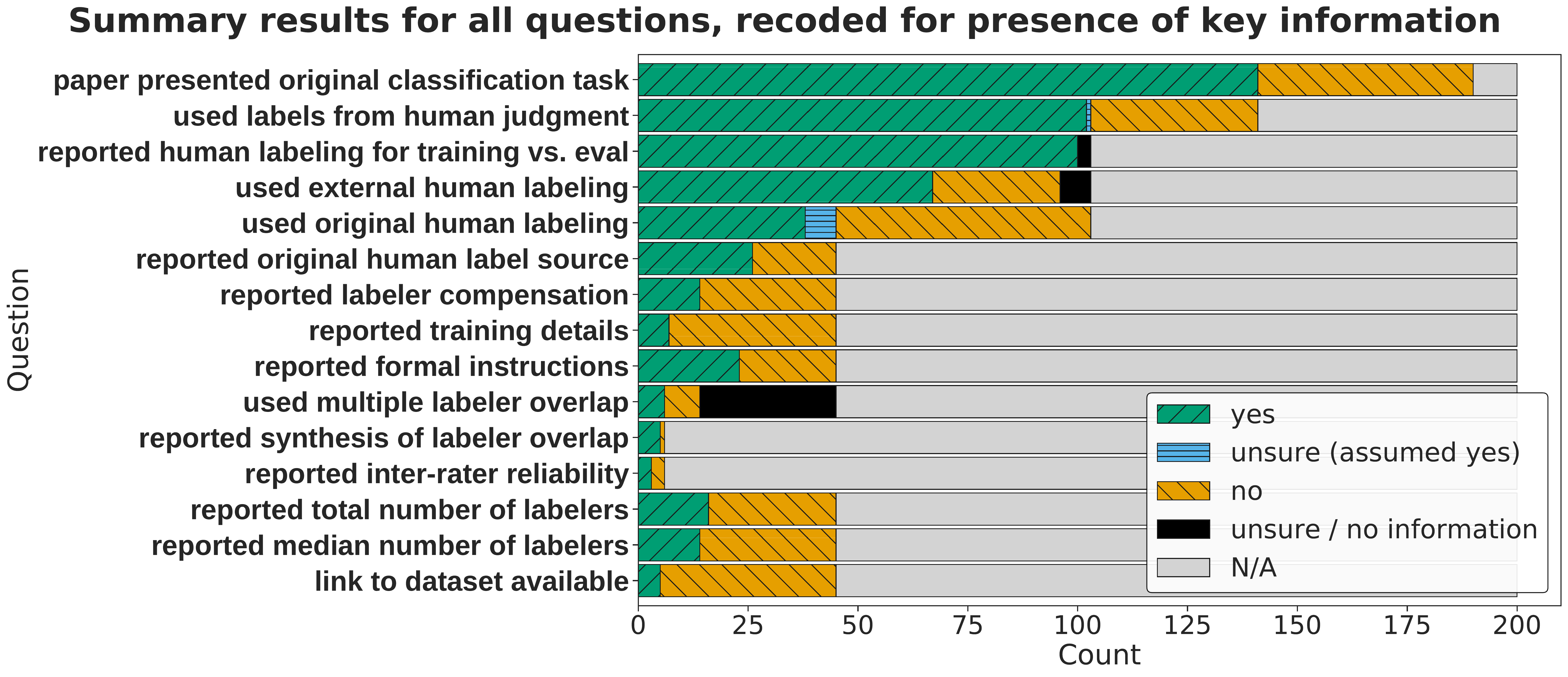}
    \caption{Summary of results. Note that some questions have been recoded to show the presence or absence of information.}
    \label{fig:info_score_summary}
\end{fullwidth}
\end{figure}
Figure \ref{fig:info_score_summary} shows a summary of results. For this figure, we recoded (or consolidated) some questions with many answers to reflect whether the paper reported an answer to that question. For example, for ``original human labeling source,'' any answer that specified a source is ``yes,'' while ''no information'' is ``no.'' This is also how we calculated paper information scores in section \ref{sec:infoscores}. Figure \ref{fig:info_score_summary} illustrates how we asked more detailed questions for papers based on answers to prior questions. For example, 103 papers used labels from human judgment --- either ''yes'' or ''unsure (assumed yes)'' --- and the next three questions were answered for those 103 papers. The remaining 10 questions were answered for the 45 papers that could be assumed to use original human labeling, with 2 of those questions only answered for the 6 papers involving multiple labeler overlap.

\setlength{\abovecaptionskip}{2pt plus 3pt minus 2pt}

\subsection{Original ML classification task}

The first question was whether the paper was conducting an original classification task using supervised machine learning. Our keyword-based process of generating the corpus included some papers that used ML keywords but were not actually presenting a new ML classifier. However, defining the boundaries of supervised ML and classification tasks is difficult, particularly for papers that are long, complex, and ambiguously worded. We defined machine learning broadly: any automated process that does not exclusively rely on explicit rules, in which the performance of a task increases with additional data \citep[][p.2]{mitchell1997}. We decided this can include simple linear regressions, although there is much debate about if and when simple linear regressions are a form of ML. However, as we were also looking for classification tasks, linear regressions were only included if it is used to make a prediction in a set of defined classes. We defined an ``original'' classifier to mean a classifier the authors made based on new or old data, which excludes the exclusive use of pre-trained classifiers or models. We found that some papers claimed to be using ML, but when we examined the details, these did not fall under our definition. 

\begin{table}[h]
\caption{Is the paper presenting a original/newly-created ML classifier?}
\begin{tabular}{lll}
\toprule
{} & Count & Proportion \\
\midrule
Yes    &   141 &     70.50\% \\
No     &    49 &     24.50\% \\
N/A (paper ineligible or inaccessible)    &    10 &      5.00\% \\
\midrule
Total  &   200 &    100.00\% \\
\bottomrule
\end{tabular}
\label{table-original-classification-task}
\end{table}

As table \ref{table-original-classification-task} shows, the majority of papers in our dataset were involved in an original classification task. We placed 10 papers in the ``N/A'' category --- meaning they did not give enough detail for us to determine, were not in English, were not able to be accessed, or were complex boundary cases.

\subsection{Classifier area/domain}

The next question categorized the paper into one of eight fields/areas of study. We had sampled three broad disciplinary categories (Social Sciences \& Humanities, Biomedical \& Life Sciences, and Physical \& Environmental Sciences), which are determined by Scopus on a per-journal/conference level. We made these area/domain determinations based on the paper's content, without consulting the Scopus-provided category. As table \ref{table-classification-outcome} shows, our data set contained a wide variety of ML application fields. Medical papers had the plurality of responses, followed by Linguistic, then papers from Biological, Physical, Soft/hardware, and Geo/ecological had similar sizes. 

\begin{table}[h!]
\caption{Classifier area/domain}
\begin{tabular}{lll}
\toprule
{} & Count & Proportion \\
\midrule
Medical               &  43 &    30.50\% \\
Linguistic            &  24 &    17.02\% \\
Biological (non-medical)            &  17 &    12.06\% \\
Physical              &  14 &     9.93\% \\
Soft/hardware         &  14 &     9.93\% \\
Geo/ecological        &  13 &     9.22\% \\
Activities and actions&   7 &     4.96\% \\
Demographic           &   5 &     3.55\% \\
Other                 &   4 &     2.84\% \\
\midrule
Total of applicable papers (presenting original ML classifier) & 141 &   100.00\% \\
\bottomrule
Non-applicable papers & 59 & -- \\
\bottomrule
\end{tabular}
\label{table-classification-outcome}
\end{table}

\subsection{Labels from human judgment}
\label{sec:labels-from-human}
While all approaches to curating training data involve some kind of human judgment, this question focused on cases where humans made discrete judgments about a set of specific items, which were then turned into labels for training data. More than a quarter of the papers in our corpora used some form of automation, scripting, or quantitative thresholds to label items. For example, one boundary case used medical records to label patients with or without high blood pressure (hypertension). We decided that if a medical practitioner made a diagnosis that researchers used as the label, it was human-labeled. If the researchers set a quantitative threshold for high blood pressure, then parsed medical records for blood pressure readings with a script, it was not human-labeled. In addition, individual human labeling could be done for all of the paper's training data (the typical case) or only a portion. For example, some authors reported using scripts or thresholds to label some items (e.g. the `easy` cases) then labeled the remaining items manually.

In some instances, we determined the answer could be an ``implicit yes'' if ample evidence indicated a particular labeling method that most likely used humans at some point, but it was not explicitly stated by the authors. For example, many medical papers reported using diagnoses from a patient's medical records as labels. Some of these papers gave substantial detail about who originally made the diagnosis and even what diagnostic criteria were used, while others generated labels based on medical records and did not explicitly state that a human (e.g. a medical practitioner) made the diagnosis. If we could reasonably assume a human was involved in the original diagnosis, we generally labeled the second type of papers as ''no information (implicit yes).'' One paper was far less clear about the source of the data than other ''implicit yes'' papers, such that we labeled it 'Unsure.'' However, we included the paper in subsequent questions because felt we could answer subsequent questions about it, which re-used externally-obtained data for labeling. 

As table \ref{table-labels-from-human-annotation} shows, the second highest response are papers that do not clearly state whether their labeling was performed by a human or a machine, but contained enough contextual details for us to be reasonably confident in assuming that human labeling was used. Note that this question was originally titled "Labels from human annotation" throughout the labeling and reconciliation process, but was renamed in the analysis stage to better reflect the instructions.

\begin{table}[h]
\caption{Were labels derived from humans making discrete judgments of items?}
\begin{tabular}{lll}
\toprule
{} & Count & Proportion \\
\midrule
No/Machine-labeled       & 38 &  26.95\% \\
Yes for all items        & 53 &  37.59\% \\
Yes for some items       & 10 &   7.09\% \\
No information (implicit yes)& 39 &  27.66\% \\
Unsure (but assumed yes)                   &  1 &   0.71\% \\
\midrule
Subtotal: papers assumed to use human labeled-data & 103 & 73.05\% \\
\midrule
Total of applicable papers (presenting original ML classifier) & 141 & 100.00\% \\
\midrule
Non-applicable papers & 59 & -- \\
\bottomrule
\end{tabular}
\label{table-labels-from-human-annotation}
\end{table}

\subsection{Human labeling for training versus evaluation}
This question and all subsequent questions were only applicable to papers that involved human labeling, which had "yes" or "implicit" designations to the previous question. This allowed for further specification of human labeled data usage within each publication. Human labeling for training data is the typical case, where labels are created and then used to train the classifier. Often part of this data is held out as a test set to evaluate the classifier. Human labeling for evaluation only is when the authors of the paper train the classifier using non-human-labeled data, but use humans to either evaluate the validity of that dataset or the classifier. The overwhelming majority of papers took the more standard approach of using labels as training data, but a few did have human evaluation of classifiers trained with machine-labeled data. This question had lower rates of ``unsure,'' where the paper did not give enough information to make a determination.

\begin{table}[h]
\caption{Was human-labeled data used for training data or to evaluate a classifier trained on non-human-labeled training data?}
\begin{tabular}{lll}
\toprule
{} & Count & Proportion \\
\midrule
Human labeling for training data         &    94  &  91.26\% \\
Human labeling for evaluation only       &     6  &   5.83\% \\
Unsure                       &     3  &   2.91\% \\
\midrule
Total of applicable papers (assumed to use human-labeled training data) & 103 & 100.00\% \\
\midrule
Non-applicable papers & 97 & -- \\
\bottomrule
\end{tabular}
\label{table-human-annotation-for-training-vs-evaluation}
\end{table}

\subsection{Original and/or external human-labeled data}
\label{sec:original_human_annot}

Our next question was about whether papers that used human labeling used original human labeling, which we defined as a process in which the paper's authors obtained new labels from human judgments for items. This is in contrast to externally-obtained data, which involves re-using existing private or public datasets of human judgments. Most of the papers in our corpus that used labels from human judgment were re-using externally-labeled data. Our assumption behind this question is that papers which rely on existing datasets may have less of a burden to discuss the details around the labeling process in the paper itself, as readers could review the cited paper for such details. In some cases, external and original human labeling were combined, such as if authors re-used a existing labeled dataset and then further labeled it for additional information. 

Like the prior question, this question had lower rates of ``unsure / no information'' where the paper did not give enough information to make a determination. We note that for all of the papers we labeled as ''unsure / no information'' we had enough contextual or implicit information to assume that it was not a re-used / externally-labeled dataset. This means that the total number of papers we assume to include at least some original human labeling is 45.

\vspace{15px}

\begin{table}[h]
\caption{Did authors re-use an existing human-labeled dataset (external), create a new human-labeled dataset (original), or both?}
\begin{tabular}{lll}
\toprule
{} & Count & Proportion \\
\midrule
Only external          &    58  &     56.31\% \\
Only original          &    29  &     28.16\% \\
Original and external  &     9  &      8.74\% \\
Unsure / no information (but can assume original)  &     7  &      6.80\% \\
\midrule
Subtotal: assumed to include some original human labeling & 45 & 43.69\% \\
\midrule
Total of applicable papers (assumed to use human-labeled training data)   &    103 &    100.00\% \\
\midrule
Non-applicable papers &   97 & -- \\
\bottomrule
\end{tabular}
\label{table-used-original-human-annotation}
\end{table}

\subsection{Summary of ML papers' approaches to training data}

\begin{figure}[t!]
    \centering
    \includegraphics[width=\textwidth]{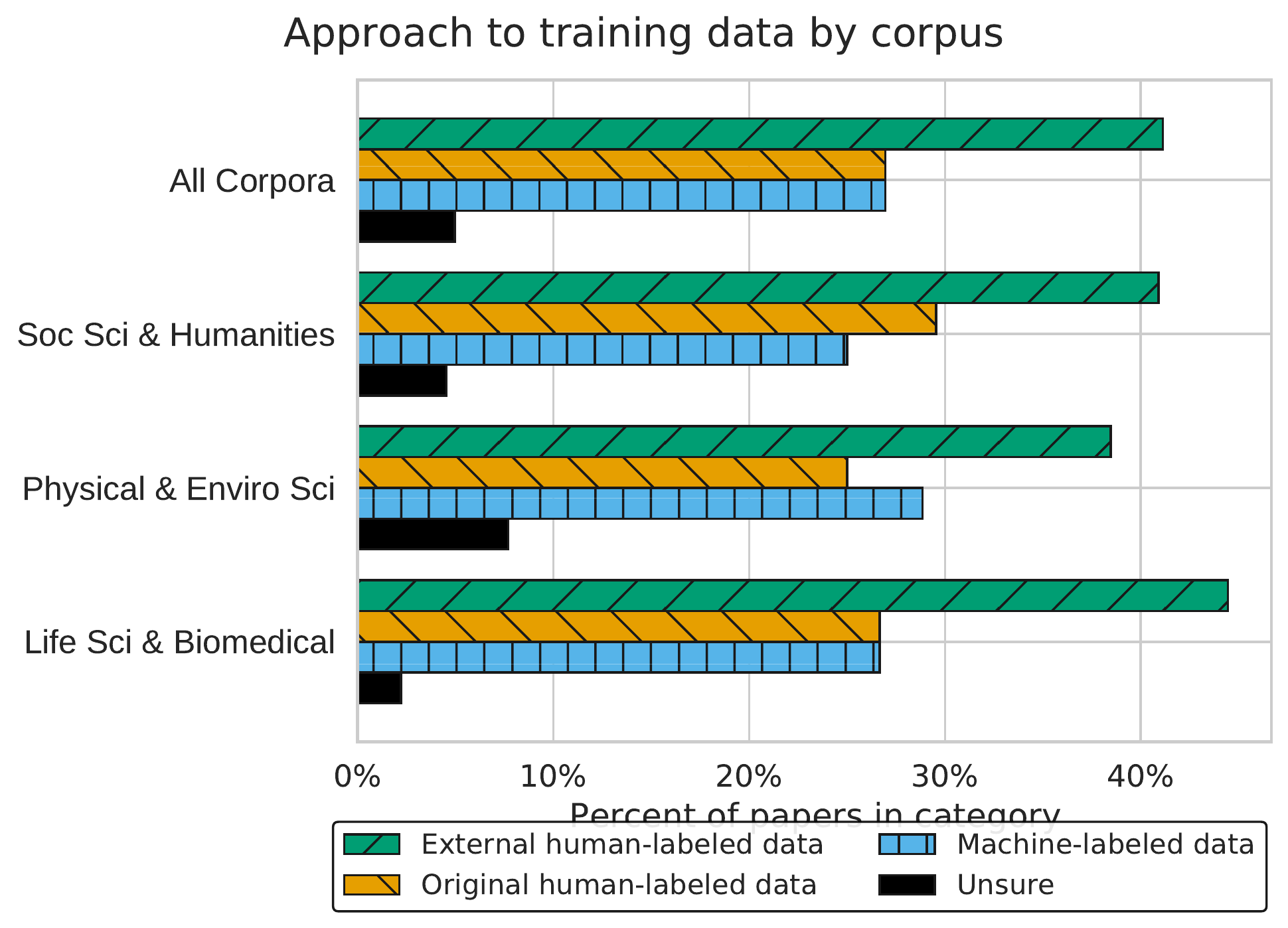}
    \caption{Approach to training data by corpus, excluding ineligible papers}
    \label{fig:approach_by_corpus_no_na}
\end{figure}

We synthesized responses to the prior questions to summarize the general breakdown of applied ML publications' approach to their data. Out of the 141 papers in our sample that presented an original ML classifier, 27\% used machine-labeled data (either machine-labeled by the authors or from a re-used dataset), 41\% used an existing human-labeled dataset, 27\% produced a novel human-labeled dataset, and 5\% did not provide enough information for us to answer. Table \ref{tab:approach_to_data} and Figure \ref{fig:approach_by_corpus_no_na} present these results by corpus, which show few differences at this level.
\begin{fullwidth}
\begin{table}[t!]
\caption{Approach to training data by corpus: count (proportion). Totals may not equal 100\% due to rounding.}
\label{tab:approach_to_data}
\begin{tabular}{|p{4.6cm}|p{1.8cm}|p{1.8cm}|p{1.85cm}|p{2cm}|}
\hline
 & \textbf{Life Sci \& Biomedical} & \textbf{Physical \& Enviro Sci} & \textbf{Soc Sci \& Humanities} & \textbf{All Corpora} \\ \hline
\textbf{Original human-labeled data} & 12 (26.7\%) & 13 (25.0\%) & 13 (29.5\%) & \textbf{38 (26.95\%)} \\ \hline
\textbf{External human-labeled data} & 20 (44.4\%) & 20 (38.5\%) & 18 (40.9\%) & \textbf{58 (41.1\%)} \\ \hline
\textbf{Machine-labeled data} & 12 (26.7\%) & 15 (28.8\%) & 11 (25.0\%) & \textbf{38 (26.95\%)} \\ \hline
\textbf{Unsure} & 1 (2.2\%) & 4 (7.7\%) & 2 (4.5\%) & \textbf{7 (5.0\%)} \\ \hline
\textit{Subtotal: ML classifier papers} & \textit{45 (100\%)} & \textit{52 (100\%)} & \textit{34 (100\%)} & \textit{\textbf{141 (100\%)}} \\ \hline
\textit{(No ML classifier / NA)} & \textit{15} & \textit{18} & \textit{26} & \textit{\textbf{59}} \\ \hline
\textbf{Grand total} & \textbf{60} & \textbf{70} & \textbf{60} & \textbf{200} \\ \hline
\end{tabular}
\end{table}
\end{fullwidth} 

\subsection{Original human labeling source}
Our next question asked who the labelers were, for the 45 papers that used original human labeling. As table \ref{table-original-human-annotation-source} shows, we found a diversity of approaches to the recruitment of human labelers.  The plurality of papers gave no information about who performed their labeling task. The ``survey/self-reported'' category refers to papers that have individuals label data they generated, which included surveys as well as studies like those using motion tracking, where subjects recorded performing different physical gestures. In contrast to Geiger et al's prior findings about papers that used Twitter data, none of the papers in our dataset reported using crowdworking platforms. We did not consider volunteer citizen science crowdsourcing platforms to be crowdworking.

\begin{table}[h]
\caption{Who were the humans doing the labeling work?}
\begin{tabular}{lll}
\toprule
{} & Count & Proportion \\
\midrule
Paper's authors             &    10 &     22.22\% \\
No information              &    19 &     42.22\% \\
Other w/ claim of expertise &     9 &     20.00\% \\
Other no claim of expertise &     2 &      4.44\% \\
Survey/self-reported        &     5 &     11.11\% \\
\midrule
Total of applicable papers (involving original human labeling) & 45 & 100.00\% \\
\midrule
Non-applicable papers & 155 & -- \\
\bottomrule
\end{tabular}
\label{table-original-human-annotation-source}
\end{table} 

\vspace{-.25cm}

\subsection{Labeler compensation}
The next question inquired as to if and what type of compensation was offered to labelers for their work. Our labels for compensation included money or gift cards, class credit, paper authorship, other compensation, explicitly stating no compensation was given (or volunteers), and no information. We observed that most publications did not provide this information, and therefore the label of "no information" was given to the majority of papers for this question. 
\begin{table}[h]
\caption{How were labelers compensated, if at all?}
\begin{tabular}{lll}
\toprule
{} & Count & Proportion \\
\midrule
Paper authorship                  &   10 &    22.22\% \\
Volunteer/explicit no compensation&    4 &     8.89\% \\
Other compensation specified      &    0 &     0.00\% \\
No information                    &   31 &    68.89\% \\
\midrule
Total of applicable papers (involving original human labeling) & 45 & 100.00\% \\
\midrule
Non-applicable papers & 155 & -- \\
\bottomrule
\end{tabular}
\label{table-annotator-compensation}
\end{table}

\subsection{Training for human labelers \& formal instructions}
The next two questions focused on how labelers were prepared for their work. We 
defined training as practicing the labeling task with interactive feedback (e.g. being told what they got right or wrong, or being able to ask questions) prior to starting the main labeling work for the study. Formal instructions are documents or videos containing guidelines, definitions, and examples that the labelers could reference as an aid. In two cases, the paper gave enough detail for us to know that no definitions or instructions were given to labelers beyond the text of the question, but about half of papers did not give enough information to make a determination.  

\begin{table}[H]
\caption{Were any details specified about how labelers were trained?}
\begin{tabular}{lll}
\toprule
{} & Count & Proportion \\
\midrule
Some training details       &     7 &     15.56\% \\
No information              &    38 &     84.44\% \\
\midrule
Total of applicable papers (involving original human labeling) & 45 & 100.00\% \\
\midrule
Non-applicable papers & 155 & -- \\
\bottomrule
\end{tabular}
\label{table-training-for-human-annotators}
\end{table}

\begin{table}[H]
\caption{What kind of formal instructions and/or examples were given to labelers?}
\begin{tabular}{lll}
\toprule
{} & Count & Proportion \\
\midrule
Instructions w/ formal definitions or examples  &    21 &     46.67\% \\
No instructions beyond question text            &     2 &      4.44\% \\
No information                                  &    22 &     48.89\% \\
\midrule
Total of applicable papers (involving original human labeling) & 45 & 100.00\% \\
\midrule
Non-applicable papers & 155 & -- \\
\bottomrule
\end{tabular}
\label{table-formal-instructions}
\end{table}

\subsection{Multiple labeler overlap}

Our next three questions were all about using multiple labelers to review the same items. Having multiple independent labelers is typically a foundational best practice in structured content analysis, so that the integrity of the labels and the schema can be evaluated (although see \citep{McDonald2019}). For multiple labeler overlap, our definitions required papers state whether all or some of the items were labeled by multiple labelers, otherwise ``no information'' was recorded. We can reasonably assume that papers which did not mention whether multiple labelers were used for each item did not engage in this more intensive process, although we cannot be certain. As table \ref{table-multiple-annotator-overlap} shows, very few papers mentioned using multiple labelers per item, with the overwhelming majority not giving any indication.

\begin{table}[h]
\caption{Multiple labeler overlap}
\begin{tabular}{lll}
\toprule
{} & Count & Proportion \\
\midrule
No                 &    8 &     17.78\% \\
Yes for all items  &    6 &     13.33\% \\
Yes for some items &    0 &      0.00\% \\
No information     &   31 &     68.89\% \\
\midrule
Total of applicable papers (involving original human labeling) & 45 & 100.00\% \\
\midrule
Non-applicable papers & 155 & -- \\
\bottomrule
\end{tabular}
\label{table-multiple-annotator-overlap}
\end{table}

\subsection{Synthesis of labeler overlap \& reported inner-rater reliability}

The next two questions built off of the previous question, which were only answered if the paper had been given the label of ``yes for all items'' or ``yes for some items.'' For these papers that had multiple labeler overlap, we examined the method by which labeler disagreement was reconciled and whether any inter-rater reliability (IRR) or inter-annotator agreement (IAA) metric was reported. We did not record what kind of IRR/IAA metric was used, such as Cohen's kappa or Krippendorff's alpha, but many different metrics were used. We also did not record what the exact statistic was, although we did notice a wide variation in what was considered an acceptable score.

\begin{table}[h]
\caption{How were disagreements between labelers reconciled?}
\begin{tabular}{lll}
\toprule
{} & Count & Proportion \\
\midrule
Qualitative/discussion      &    3 &     50.00\% \\
Quantitative/no discussion  &    2 &     33.33\% \\
No information              &    1 &     16.67\% \\
\midrule
Total of applicable papers (involving multiple overlap) & 6 & 100.00\% \\
\midrule
Non-applicable papers & 194 & -- \\
\bottomrule
\end{tabular}
\label{table-synthesis-of-annotator-overlap}
\end{table}

\begin{table}[h]
\caption{Did the paper report an inter-rater reliability metric?}
\begin{tabular}{lll}
\toprule
{} & Count & Proportion \\
\midrule
Yes &    3 &     50.00\% \\
No  &    3 &     50.00\% \\
\midrule
Total of applicable papers (involving multiple overlap) & 6 & 100.00\% \\
\midrule
Non-applicable papers & 194 & -- \\
\bottomrule
\end{tabular}
\label{table-reported-inter-annotator-agreement}
\end{table}

\subsection{Total and median number of human labelers}
\label{sec:num_annot}
We then asked two final questions regarding how many individuals completed a paper's labeling task. Because this information can be presented differently based on the labeling process, we divided this into two. The total number of human labelers referred to all human labelers involved in the project at any time. The median number of human labelers per item referred to how many labelers evaluated each item in a publication's dataset, which were greater than one in the case of papers that had multiple labelers per item. Eight papers specified that there was only one labeler per item, which matches with the data in the first question about multiple labeler overlap. The majority of the papers did not provide enough information to answer the question.

\begin{table}[h]
\caption{Total number of labelers in the project}
\begin{tabular}{lll}
\toprule
{} & Count & Proportion \\
\midrule
1  &   2 &     4.44\% \\
2 &    6 &    13.33\% \\
3 &    2 &     4.44\% \\
5 &    1 &     2.22\% \\
7 &    1 &     2.22\% \\
10 &   1 &     2.22\% \\
30 &   2 &     4.44\% \\
659 &  1 &     2.22\% \\
??? & 29 &    64.44\% \\
\midrule
Total of applicable papers (involving original human labeling) & 45 & 100.00\% \\
\midrule
Non-applicable papers & 155 & -- \\
\bottomrule
\end{tabular}
\label{table-total-num-of-annotators}
\end{table}

\begin{table}[H]
\caption{Median number of labelers per item}
\begin{tabular}{lll}
\toprule
{} & Count & Proportion \\
\midrule
1  &    8 &     17.78\% \\
2  &    5 &     11.11\% \\
3  &    1 &      2.22\% \\
?? &   31 &     68.89\% \\
\midrule
Total of applicable papers (involving original human labeling) & 45 & 100.00\% \\
\midrule
Non-applicable papers & 155 & -- \\
\bottomrule
\end{tabular}
\label{table-median-num-of-annotators}
\end{table}

\subsection{Link to dataset available}

Our final question was about whether the paper contained a link to the dataset containing the original human-labeled training dataset. Note that this question was only answered for papers involving some kind of original or novel human labeling, and papers that were exclusively re-using an existing open or public dataset were left blank to avoid double-counting.  We did not follow such links or verify that such data was actually available. As table \ref{table-link-to-dataset-available} shows, the overwhelming majority of papers did not include such a link, with 5 papers (11.11\%) using original human-labeled training datasets linking to such data. Given the time, labor, expertise, and funding in creating original human labeled datasets, authors may be hesitant to release such data until they feel they have published as many papers as they can, especially junior scholars. Data sharing also requires specific expertise in data formats, documentation, and platforms, which may not be equally distributed across academic disciplines.

\begin{table}[H]
\caption{Link to dataset available}
\begin{tabular}{lll}
\toprule
{} & Count & Proportion \\
\midrule
No  &    40 &     88.89\% \\
Yes &     5 &     11.11\% \\
\midrule
Total of applicable papers (involving original human labeling) & 45 & 100.00\% \\
\midrule
Non-applicable papers & 155 & -- \\
\bottomrule
\end{tabular}
\label{table-link-to-dataset-available}
\end{table}

\section{Paper information scores}
\label{sec:infoscores}
After finalizing the labels, we quantified the information each paper provided about training data, based on how many questions we could answer for each paper. We developed a total and normalized information score, as different studies demanded different levels of information. For example, our questions about whether inter-rater reliability metrics and reconciliation methods were reported are only applicable for papers involving multiple labelers per item. However, all other questions are relevant for any project involving original human labeling. As such, papers involving original human labeling without multiple labelers per item had a maximum of 11 points, while those with multiple labelers per item had a maximum of 13 points. The normalized score is the total score divided by the maximum score possible.

\subsection{Overall distributions of information scores}

\begin{figure}[h!]
    \centering
    \includegraphics[width=\textwidth]{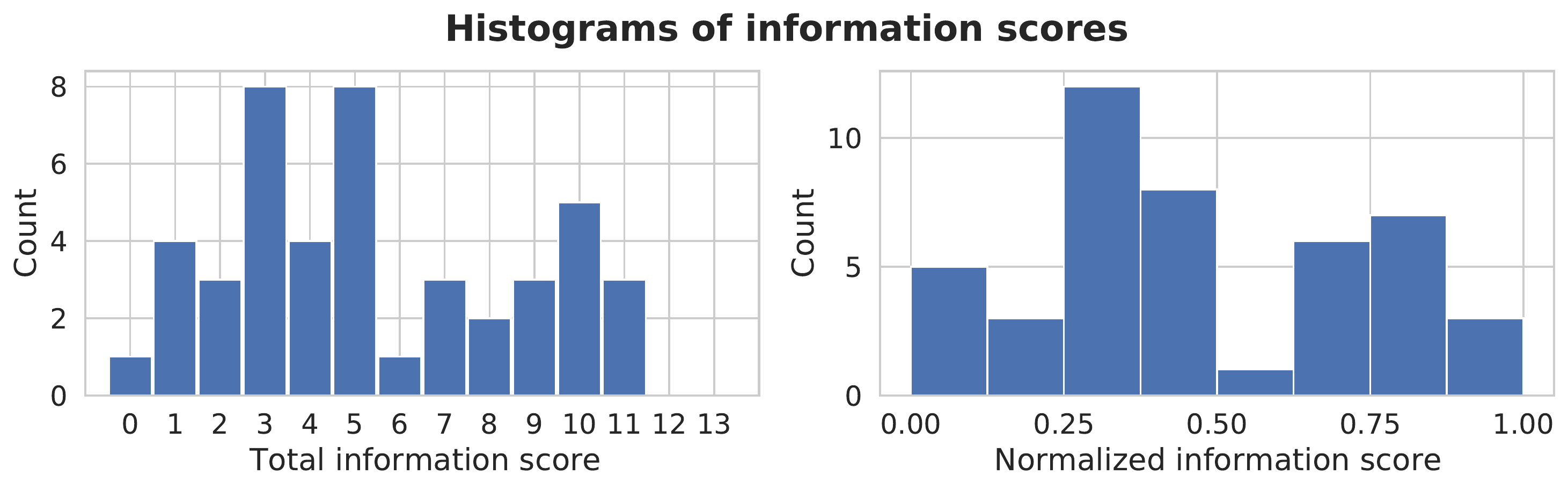}
    \caption{Histograms of total and normalized information scores for all papers involving original human labeling.}
    \label{fig:info_score_raw_norm_hist}
\end{figure}

Figure \ref{fig:info_score_raw_norm_hist} shows histograms for total and normalized information scores, which show that scores varied substantially. As Geiger et al \citeyearpar{geiger2020garbage} also found, this roughly suggests two overlapping distributions and thus populations of publications: one centered around total scores of 3-5 and normalized scores of 0.3 and another centered around total scores of 9 and normalized scores of 0.7. The normalized information score ranged from 0 to 1, with 1 paper having a normalized score of 0 and 3 papers with a full score of 1. The total information score ranged from 0 to 11, with no paper receiving a full score of 13, which would have required a study involving multiple labeler overlap that gave answers to all questions, including IRR metrics and reconciliation method. Overall, the mean total score was 5.4, with a median of 5 and a standard deviation of 3.2. The mean normalized information score was 0.472, with a median of 0.455 and a standard deviation of 0.268. This is quite similar to the findings by Geiger et al \citeyearpar{geiger2020garbage} for their normalized scores, which had a mean of 0.441, a median of 0.429, and a standard deviation of 0.261.

\subsection{Information scores by corpus and application areas}

\begin{figure}[h!]

    \centering
    \includegraphics[width=.95\textwidth]{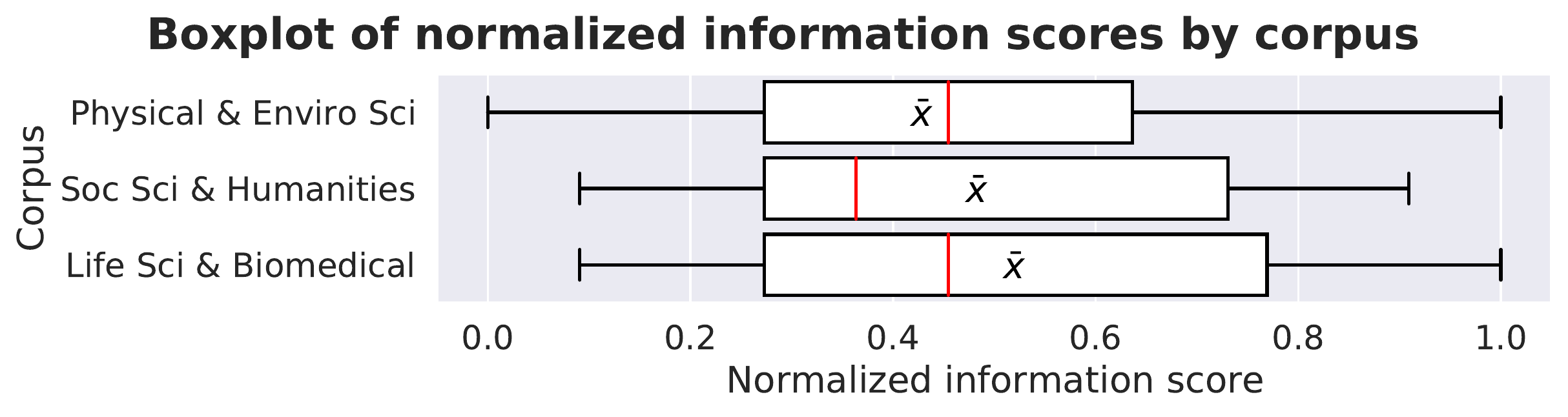}
    \caption{Boxplots of normalized information scores for papers using original human labeling, by corpus}
    \label{fig:info_score_boxplot_corpus}
\end{figure}

We analyzed information scores by corpus for all papers using original human labeling. Figure \ref{fig:info_score_boxplot_corpus} is a boxplot illustrating the distribution of normalized information scores by corpus.\footnote{For this and all other boxplots in this paper: The main box is the inter-quartile range (IQR), or the 25th \& 75th percentiles. The middle red line is the median, the black $\overline{X}$ is the mean. The outer whiskers are the highest and lowest data points in a range of 1.5 times the IQR from the median. Grey diamonds are outliers beyond 1.5 times the IQR from the median.}  There was a lower median score (red lines in boxplots) for social science \& humanities papers (0.364) than life science \& biomedical papers (0.455) and physical \& ecological science papers (0.455). However, when examining means between groups ($\overline{X}$ in boxplots), the physical \& ecological science papers had a lower mean (0.428) than social science \& humanities papers (0.482) and life science \& biomedical papers (0.519). We ran a 1-way analysis of variance (ANOVA) of normalized information scores by corpus. No statistically significant difference was found ($p=0.65$, $F=0.43$). Because we run 3 statistical tests in this paper, we apply a Bonferroni correction to address the multiple comparisons problem \citep{dunn1961multiple}, moving our p-value target from 0.05 to 0.0166. 

\begin{figure}[h]
    \centering
    \includegraphics[width=.97\textwidth]{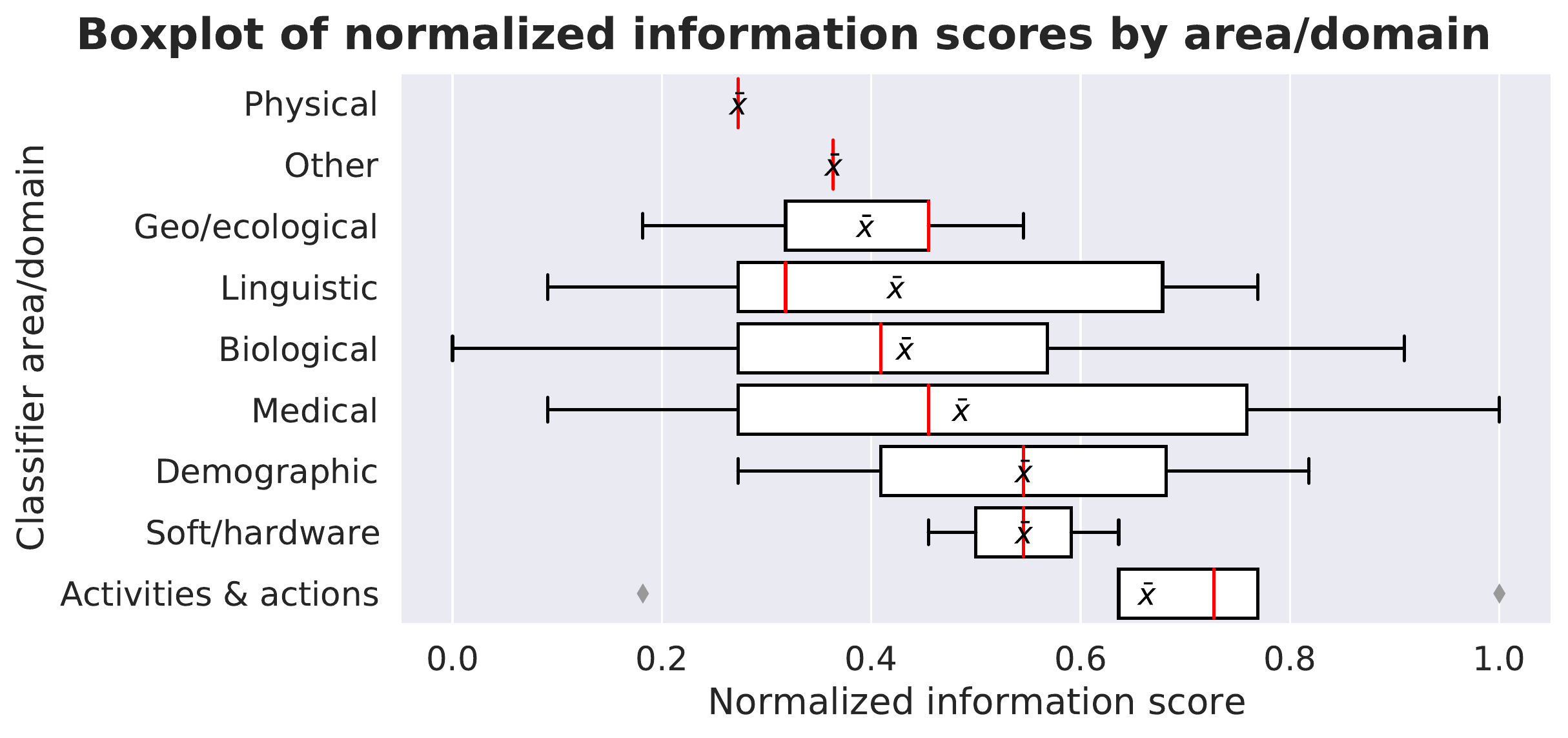}
    \caption{Boxplots of normalized information scores for all papers involving original human labeling by application area/domain. Physical and Other only had 1 paper. Activities \& actions do not have whiskers because no items had scores from 1 to 1.5 * IQR, but did have two outliers outside the 1.5 * IQR range.  }
    \label{fig:info_score_boxplot_area}
\end{figure}

Next, we conducted a similar analysis by the classification area/domain. A boxplot showing the distribution of normalized information scores is shown in Figure \ref{fig:info_score_boxplot_area}. These were not stratified random samples, and we ended up with far more papers in some categories than others, with only 1 item for Physical and Other. The small sample size makes formal statistical tests difficult to interpret, and the assumption of homoscedasticity is not fulfilled due to the wide range in standard deviations between these groups (e.g. 0.13 for Geo/ecological to 0.39 for Demographic). We recommend against making generalizable statistical tests or generalizations based on this analysis, but we report these scores to inform future work. Most groups' mean and median scores were between 0.4 and 0.6, with papers in the Linguistic category having lower medians (0.318). The most common categories --- Linguistic, Medical, and Biological --- also had much wider distributions and IQRs, but similar means. Activities \& actions was the highest scoring category in terms of the mean, median, and upper and lower IQR. In these studies, it is generally the case that the data are recordings of a person performing an activity, and each label is the activity they are asked to perform. This research design may lead authors to more concretely detail such methods.

\subsection{Normalized information scores by document type}
\begin{figure}[h]
    \centering
    \includegraphics[width=.9\textwidth]{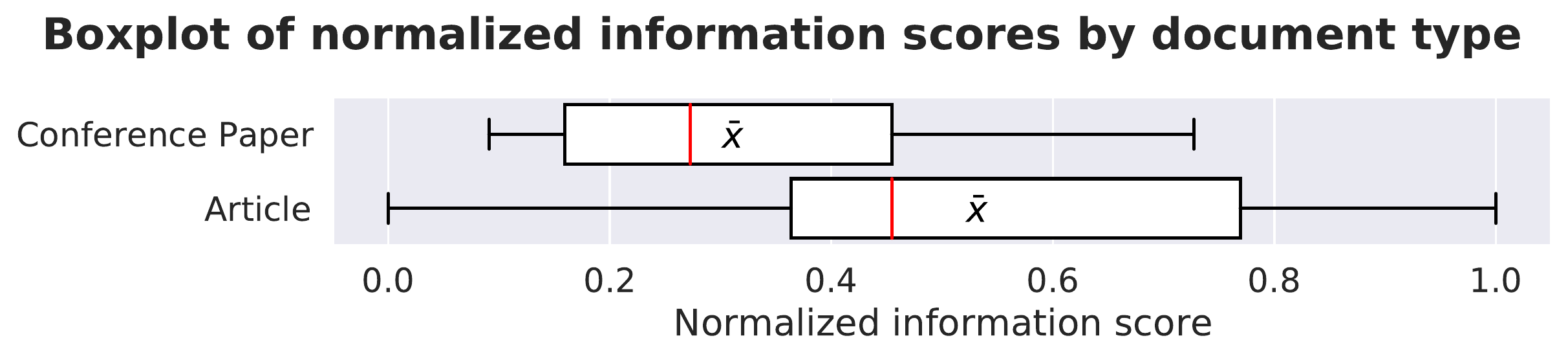}
    \caption{Boxplot of normalized information scores for papers using original labeling, by document type.}
    \label{fig:info_score_doc_type}
\end{figure}

For the 45 papers using original human labeling, 33 were journal articles and 12 were conference papers. We conducted an analysis of normalized information scores by document type, which showed larger differences. As figure \ref{fig:info_score_doc_type} shows, articles have a higher mean (0.53 vs 0.31) and median (0.45 vs 0.27). We ran a 2-tailed Welch's unequal variances t-test \citep{welch1947generalization} (variances differed by 0.024) and found a statistically significant difference ($p=0.0086$, $t=2.86$). We applied a Bonferroni correction to the p-value threshold to address the problem of multiple comparisons, but the the p-value is well below our adjusted target of 0.0166. This means that in our sample, we can assume that articles generally provide more information about training data than conference papers.

\subsection{Label source information scores}

Finally, because of the relatively small number of papers involving original human labeling (n=45) that lead to low statistical power for paper information scores, we examined all papers that presented an original ML classifier (n=141) based on whether they gave information sufficient to determine if their dataset's labels were derived from original human labeling. As discussed in section \ref{sec:labels-from-human}, we gave many papers the answer "no information (implicit yes)," which means we could reasonably assume labels were made by humans, but the paper never explicitly said humans were involved. Papers with answers "Yes for all items," "Yes for some items," and "No / machine-labeled" were scored 1. Papers with answers "No information (implicit yes)" and "Unsure (but assumed yes)" were scored 0. N/A papers that did not present an original classifier were excluded.

\begin{figure}[b!]
    \centering
    \includegraphics[width=\textwidth]{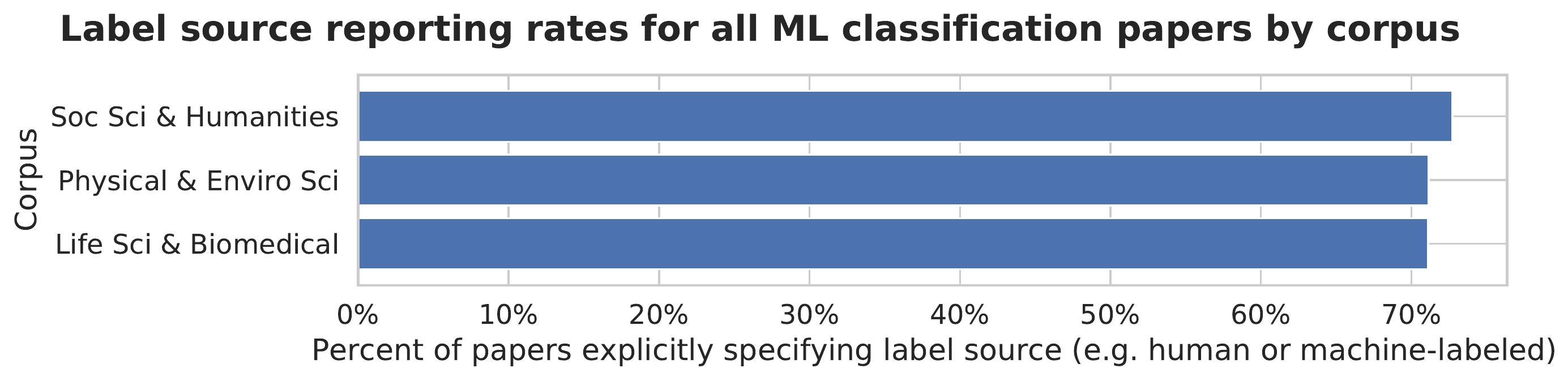}
    \caption{Label source reporting rates for papers presenting an original classifier, by corpus}
    \label{fig:source_rates_corpus}

    \centering
    \includegraphics[width=\textwidth]{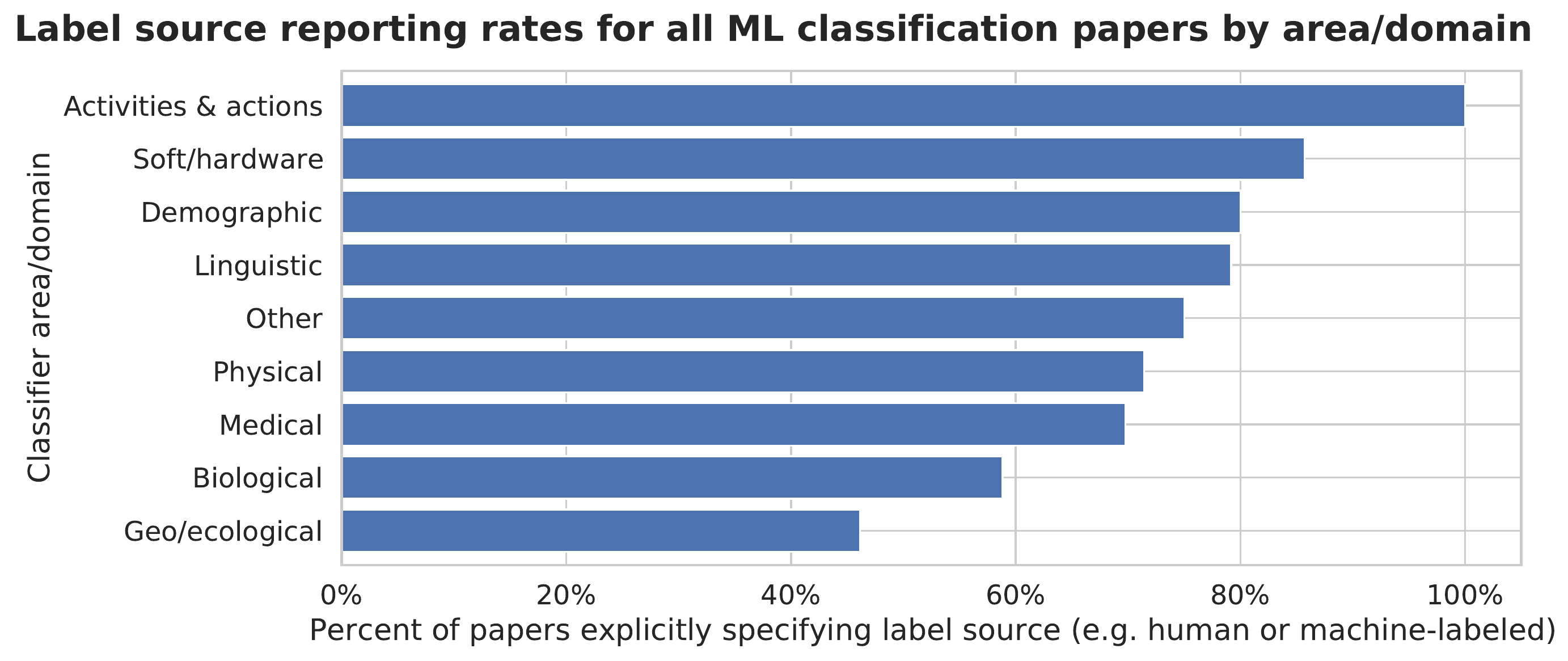}
    \caption{Label source reporting rates for papers presenting an original classifier, by area}
    \label{fig:source_rates_area}
\end{figure}

Figure \ref{fig:source_rates_corpus} shows the label source reporting rates by corpus, which shows strikingly similar rates. Social Science \& Humanities papers had a rate of 72.7\%, compared to rates of 71.1\% for the other two corpora. Figure \ref{fig:source_rates_area} shows the label source reporting rates by application area, which shows a much wider range. Activities \& actions also has the highest rate at 100\% (likely for the same reasons hypothesized earlier), with the lowest rate being Geo/ecological at 46.1\%. We also note the differences in these results and the overall paper information scores, which were inversely-ranked for the larger categories of Linguistic, Medical, and Biological. While Lingusitic papers had lower median information scores, they had far higher rates of label source reporting (79.2\%), compared to Medical (69.8\%) and especially Biological (58.8\%) papers. 

\begin{figure}
    \centering
    \includegraphics[width=\textwidth]{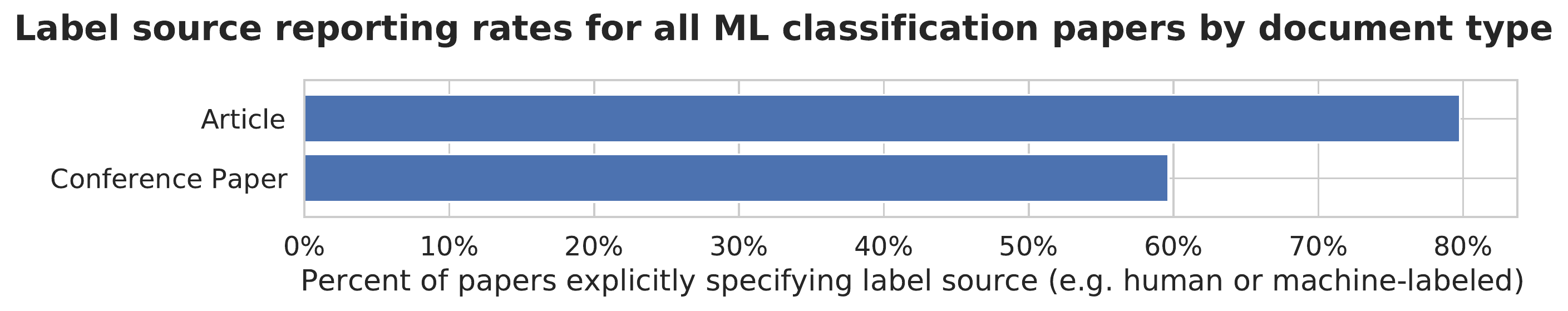}
    \caption{Label source reporting rates for papers presenting an original classifier, by document type}
    \label{fig:source_rates_doc_type}
\end{figure}
Figure \ref{fig:source_rates_doc_type} shows label source reporting rates by journal article versus conference paper, which shows a much higher rate for articles. We ran a 2-tailed Welch's unequal variances t-test (variances differed by 0.082) and a statistically significant difference was not found ($p=0.038$, $t=2.35$). We must apply a Bonferroni correction to the p-value threshold to address the problem of multiple comparisons, and the the p-value is above our adjusted target of 0.0166.

\subsection{Conclusion to information score results}
In conclusion, our quantitative metrics show quite varying ranges and distributions of information scores, which does give evidence to the claim that there is substantial and wide variation in the practices around human labeling, training data curation, and research documentation. The ranges of the boxplots of normalized information scores are substantial, both for IQRs (25th \& 75th percentile) and the whiskers at 1.5 * IQRs. Ranges are larger when sampling by corpus, but still substantial for the application areas with more papers (e.g. Medical, Biological, Linguistic). 

We specifically call for more investigation into applied ML geo/ecological research, which often classifies land use from aerial photos or photos of geological samples. These had the lowest rates of label source specification and the lowest mean normalized information scores (excluding the categories that only had 1 paper). However, from our experience, some papers with lower scores did give excellent levels of detail about how they were using an existing land use classification schema they cited (e.g. the widely-used USGS guide by \cite{anderson1976land}), but did not give any details about who applied that schema to the aerial photos. We can also hypothesize that in fields with widely-established and shared methodological standards, researchers could have far higher rates of adherence to methodological best practices around data labeling, but have lower rates of reporting that they actually followed those practices in papers.

Finally, we draw attention to the different rates when we grouped by corpus versus application area. In our sampling, the corpus was the Scopus-provided metadata field, which is determined at the publication level when a journal is added to Scopus.\footnote{\url{https://web.archive.org/web/20210531200329/https://service.elsevier.com/app/answers/detail/a_id/14882/supporthub/}} At this level, we saw fewer differences in quantitative scores. In contrast, our application area field is determined for each paper based on the content, independent of the journal or venue in which it was published. Scores varied far more when broken out by application area, which is likely due in part to noise in the smaller samples. However, this may also indicate that methodological reporting rates vary even more within sub-fields or types of research objects. For future work, we recommend that researchers pay specific attention to differences between fields or objects of study, rather than solely group papers in the high-level way we did with our three corpora.

\section{Concluding discussion}
\subsection{Findings}

First, our study shows that contemporary applications of supervised machine learning across disciplines often rely on training datasets in ways that either reuse existing human-labeled datasets or label items with some kind of automated process. Of the papers in our dataset that presented an original ML classifier, only 26.7\% produced a new human-labeled dataset as part of their study --- a rate that did not substantially vary among our three corpora from the biomedical \& life sciences, the physical \& environmental sciences, and the humanities \& social sciences. Second, of the applied ML publications that did produce a new human-labeled training dataset, there was significant divergence in reporting methodological details and following best practices in human labeling. A small number of publications received top information scores, but approximately two-thirds of publications involving original human labeling did not provide enough information for us to answer more than half of the subsequent questions we asked about the labeling process. 

This cross-disciplinary trend is cause for concern, given that high-quality training data is essential to the validity of machine learning classifiers and human judgment is notoriously difficult to standardize. When comparing across our three broad corpora of social science \& humanities, biomedical \& life sciences, and physical \& environmental sciences papers, we only see marginal differences in the level of information papers provide. We do see more robust evidence that journal articles have higher rates of reporting information about training data than conference papers, which may relate to conference papers being shorter and only involving a single cycle of peer-review.\footnote{Not all conference papers are peer-reviewed, but all conference proceedings indexed by Scopus are be peer-reviewed.}

\subsection{Implications}

\subsubsection{The black-boxing of training data}

Machine learning is increasingly used across disciplines and application domains, but the quality of supervised ML classifiers is only as good as the data that is used to train it. Based on our findings, we argue for more attention to be placed on the specific details of how that training data is labeled. There is a recent wave of work that interrogates ML models once they are trained, as well as considerations about ``automation bias'' \citep{skitka1999does} --- that people often treat trained models as a ``black box,'' with their outputs unquestioned and taken as given. These concerns must also extend to the labeling and curation of training datasets, some of which become widely re-used without being examined. 

For example, Crawford and Paglen \citeyearpar{excavatingai} have called attention to problematic racial labels of images in the popular ImageNet training dataset, which has been a standard benchmark dataset in image recognition for over a decade. Birhane and Prabhu \citeyearpar{Birhane_2021_WACV} found thousands of images in the 80 Million Tiny Images dataset that were labeled with offensive racial and gender-based slurs. The careful curation of datasets has long been a central tenet in the institutions of science, although standards and practices can change dramatically over time and across contexts. Historians of science like Bowker \citeyearpar{bowker_memory_2005} and Gitelman \citeyearpar{gitelman2013raw} remind us that data is never "raw," as data always is produced and used within a messy assemblage of partially-overlapping human institutions, each of which have their own practices, values, and assumptions. To this end, we call for applied ML researchers and practitioners who are re-using human-labeled datasets to exercise as much caution and care around the decisions to re-use a labeled dataset as they would if they were labeling the data themselves. 

Finally, we have not asked any questions about how papers discuss data cleaning, but we encourage more investigation and consideration of how the often-backgrounded work of data cleaning is performed, managed, and documented. We could have asked another dozen questions about how papers did or did not discuss how they cleaned their data. For future work, we would encourage researchers to study what applied ML papers report about how they cleaned and pre-processed their data. We also see much future work in studying to what extent applied ML papers report efforts at de-biasing datasets and models.

\subsubsection{Institutional change around data documentation}
We call on the institutions of science --- publications, funders, disciplinary societies, and educators --- to play a major role in working out solutions to these issues of data quality and research documentation. We see this work as part of the open science and reproducibility movement, specifically the movement for open access to research datasets, materials, protocols, and analysis code. However, even advocates of this movement have long discussed how individual researchers do not have incentives to be first-movers in being more open than usual about the messiness in all research, because it leaves their work more open to rebuttal \citep{smaldino_why_2016,ali_motivating_2017,zimring_were_2019}. In our own experience, we have certainly felt the temptation to not report certain details that would lead others to have less confidence in our study, such as our inter-rater reliability metrics. 

In looking towards solutions, we see a parallel to issues in open access to publications, which often requires individual researchers to choose if they want to pay for open access out of their own funding. While some first-movers paid for this out of their own budgets, open access is currently being far more effectively tackled at the institutional level in ways that will not require individuals (and especially first-movers) to pay the costs. So too do we see institutional solutions to the issue of methodological detail, where a common floor could be established that is equally applicable to all researchers. We also see resonance with the various proposed efforts at standardizing documentation about machine learning models and datasets \citep{gebru2018datasheets,mitchell2019model,bender2018data,holland2018dataset,barclay2019towards,beretta2018ethical,hind2018increasing,raji_about_2019} and urge that human labeling details be included in such efforts.

On the publication process, we note that research publications are limited by length restrictions, which can leave little space for details. We can hypothesize that having a dedicated and visible space for methodology and dataset documentation would make these concerns more central for authors, reviewers, editors, and readers, although we can only speculate as to the best way for this to be implemented. For example, \textit{Nature} has far shorter word limits for a main research article (2,000 to 2,500 words), which means methodological and dataset documentation is often fully detailed in appendices, which can be of any length. Does this approach more easily lead to readers and/or reviewers ignoring such details and focusing more on results? However, \textit{Nature} also requires that authors fill out a peer-reviewed checklist form that asks general and domain-specific questions about statistical details (e.g. ``a description of all covariates tested``) and about the dataset (e.g. for behavioral science, ``State the research sample ... provide relevant demographic information ... and indicate whether the sample is representative'').\footnote{\url{https://www.nature.com/documents/nr-reporting-summary-flat.pdf}} Do these kinds of mandatory structured disclosure forms make these concerns more central to authors and reviewers, even if they are not as accessible to readers?

We also note that peer reviewers and editors play a major role in deciding what details are considered extraneous. First, we urge reviewers to make space for what some may see as ``boring'' methodological details. More importantly, we call on editorial boards to openly signal in author and reviewer guidelines that they invite or even require extended discussion of methodological details. To this end, one recent trend is the growth of multi-stakeholder groups that have collectively released formal guidelines or best practices statements on research reporting, such as the CONSORT guidelines for reporting randomized clinical trials \citep{schulz_consort_2010}, the COREQ guidelines for reporting qualitative research \citep{Tong2007}, or the PRISMA guidelines on reporting meta-analyses and systematic reviews \citep{moher2009preferred}. 

For example, PRISMA guidelines on reporting meta-analyses and systematic reviews have been mandated in the author guidelines of many journals (including \textit{The Lancet},\footnote{\url{https://els-jbs-prod-cdn.jbs.elsevierhealth.com/pb/assets/raw/Lancet/authors/tl-info-for-authors.pdf}} \textit{PLoS ONE},\footnote{\url{https://journals.plos.org/plosone/s/submission-guidelines\#loc-systematic-reviews-and-meta-analyses}} and \textit{Systematic Reviews}\footnote{\url{https://systematicreviewsjournal.biomedcentral.com/submission-guidelines/preparing-your-manuscript/research}}), which require authors fill out the 27-item PRISMA checklist.\footnote{\url{http://prisma-statement.org/documents/PRISMA\%202009\%20checklist.pdf}} One interesting trend with such multi-stakeholder best practices statements in medicine is the proliferation of subdomain-specific ``extensions'' that further specify methodological reporting standards. For example, the EQUATOR network tracks 32 extensions to the CONSORT guidelines,\footnote{\url{https://www.equator-network.org/?post_type=eq_guidelines\&eq_guidelines_clinical_specialty=0\&s=+CONSORT+extension}} including guidelines for reporting randomized clinical trials in pain management \citep{gewandter_checklist_2019}, orthodontics \citep{pandis_consort_2015}, and psycho-social interventions \citep{montgomery_reporting_2018}.

However, there have been disagreements over the impact and efficacy of these more structured approaches. Page and Moher's\footnote{Moher is the lead author of the PRISMA statement.} meta-analysis of 57 papers studying uptake of the PRISMA guidelines \citep{page_evaluations_2017} found that while more papers are reporting details in the PRISMA guidelines after it was released in 2009, some details remain low even for papers claiming to adhere to the guidelines. For example, for 9 of the PRISMA items, fewer than 67\% of papers actually reported the information in question. Fleming et al.  \citeyearpar{fleming_blinded_2014} found that following the widespread uptake of the PRISMA guidelines by certain publications, more meta-analysis articles reported methodological details, but disproportionately those in the PRISMA guidelines. The authors of that study raise concerns that PRISMA has overdetermined the peer review process: authors who are fully-compliant with PRISMA are no longer reporting other methodological details that Fleming et al. claim are also relevant in such work and were in other competing meta-analysis guidelines that ultimately lost to PRISMA.

\subsubsection{Are there universal best practices for the labeling of training data?}
The efforts around methodological standards in medicine raise an important question about the wisdom of seeking a single one-size-fits-all set of best practices for any application of supervised ML.  However, contemporary efforts around ``fairness'' or ``transparency'' in machine learning often work towards more universal or domain-independent approaches, which are applied to a wide range of application areas (e.g. finance, social services, policing, hiring, medicine). Yet in our work examining publications from quite different academic fields, we found ourselves needing to pay close attention to the various kinds of specialized expertise that are required to label a training dataset for a particular purpose. As Bowker and Star \citeyearpar{bowker1999sorting} and Goodwin \citeyearpar{Goodwin1994} discuss, all classification systems rely on a shared cultural context, which can be exceedingly difficult to formally specify and often falls apart at the edges. It can be difficult to know beforehand what level of shared cultural context and expertise will be involved. 

Some of the papers we analyzed described in great detail how the people who labeled their dataset were chosen for their expertise, from seasoned medical practitioners diagnosing diseases to youth familiar with social media slang in multiple languages. That said, not all labeling tasks require years of specialized expertise, such as more straightforward tasks we saw, like distinguishing positive versus negative business reviews or identifying different hand gestures. Even projects in the same domain can require different levels of expertise, such as a dataset of animal photos labeled just for the presence of cats and dogs, versus labeling the same photos for the specific breed of cats and dogs. Furthermore, we found that some labeling tasks are well-suited to semi-automated labeling where labelers are assisted with rule-based approaches, while others are not. Finally, even the more seemingly-straightforward classification tasks can still have substantial room for ambiguity and error for the inevitable edge cases, which require training and verification processes to ensure a standardized dataset. 

The labeling protocol and schema we developed and used in this paper --- which is based on and extends prior work \citep{geiger2020garbage} --- is an effort at creating a cross-disciplinary standard for any given research project that uses human-labeled training data. While we believe that any peer reviewer or reader can ask these same questions of any ML application paper, they are only a starting point. Issues of validity, consistency, reliability, reproducibility, and accountability require further investigation. The kind of domain-independent criteria we used should be seen as necessary but not sufficient criteria for having confidence in a labeled dataset. We do not advocate for a single, universal, one-size-fits-all solution, but instead seek to spur conversations within and across disciplines about better approaches to bring the work of data labeling into the foreground. We see a role for the classic principle of reproducibility, but for data labeling: does the paper provide enough detail so that another researcher could hypothetically recruit a similar team of labelers, give them the same instructions and training, reconcile disagreements similarly, and have them produce a similarly-labeled dataset? 

Data publications could also play a major role in this issue, which are stand-alone peer-reviewed publications that do not answer a research question, but instead spend the entire paper describing the creation of a dataset in rich detail \citep{costello_motivating_2009,smith_data_2009,chavan_data_2011,candela_data_2015}. In seeking to bring the work of data labeling from the background to the foreground, our work is also aligned with scholars who have focused on the often under-compensated labor of crowdworkers and have called for researchers to detail how much they pay for data labeling \citep{silberman2018responsible}. 

\subsection{Limitations}
To conclude, we reflect that our study also has the same kinds of limitations that many human labeling projects have. For example, given the concerns we raise about domain-specific expertise, our team may have missed or misinterpreted crucial details when examining papers. The second issue is around the reliability and reproducibility of our team's labeling process. In conducting this study, we have become quite familiar with the difficulties of getting a medium-sized team to build a consensus around reducing complex objects into quantifiable data. We specifically chose to have a more detailed and time-intensive process in which disagreements were discussed, which traded off with the total number of items we were able to label. We believe this trade-off was the right decision, given our focus on methodological rigor, but it does mean our samples are smaller than we would like. The lower sample size means that we have less confidence in the statistical generalizability of our sample to the population of all applied ML publications. However, we see a wide range of future work that can be done to extend these efforts, such as with teams of domain-specific experts that examine applied ML fields in their area of expertise. 

Finally, we only have access to what each publication reported about the work they did, and not the research project itself, which means our unit of analysis is methodological reporting. For example, researchers could have far higher rates of following methodological best practices around data labeling, but have lower rates of reporting that they actually followed those practices in papers. We could even hypothesize an inverse relationship between a field's overall adherence to methodological best practices and researchers' rates of reporting adherence to those practices, if such practices become so routine and mundane that they are left implicit in publications. For these reasons, we strongly advise against interpreting our quantitative scores as an unproblematic proxy for methodological rigor, especially for the scores by discipline and area. However, given our interest in how labeling practices impact the validity of ML models and classifiers, future work could extend this work through other methods, such as surveys and ethnographic studies of ML researchers.

\section*{Acknowledgments}
We would like to acknowledge support for this project from the UC-Berkeley Undergraduate Research Apprenticeship Program (URAP) and to Stacey Dorton for administrative and logistical support. We thank the peer reviewers for their feedback and suggestions. 

\section*{Funding information}
This work was funded in part by the Gordon \& Betty Moore Foundation (Grant GBMF3834) and Alfred P. Sloan Foundation (Grant 2013-10-27), as part of the Moore-Sloan Data Science Environments grant to UC-Berkeley. The funders had no role in study design, data collection and interpretation, or the decision to submit the work for publication. 

\section*{Competing interests}

We declare no competing interests or conflicts of interest.

\section*{Data availability}
All datasets, analysis scripts, protocols, labeling instructions, and other supplementary information required to replicate and reproduce these findings can be downloaded on GitHub\footnote{ \url{https://github.com/staeiou/gigo_qss_2021}} and Zenodo.\footnote{ \url{https://doi.org/10.5281/zenodo.4906636}} 

\section*{Author contributions}
\begin{itemize}
    \item R. Stuart Geiger (\orcidlink{0000-0001-7215-0532} 0000-0001-7215-0532): Conceptualization, Data curation, Formal Analysis, Funding acquisition, Investigation, Methodology, Project administration, Resources, Software, Supervision, Validation, Visualization, Writing – original draft, Writing – review \& editing
    \item Dominique Cope (\orcidlink{0000-0003-2471-772X} 0000-0003-2471-772X): Investigation, Validation, Writing – original draft, Writing – review \& editing
    \item Jamie Ip (\orcidlink{0000-0001-9952-4987} 0000-0001-9952-4987): Data curation, Investigation, Software, Validation
    \item Marsha Lotosh (\orcidlink{0000-0003-0362-8947} 0000-0003-0362-8947): Investigation, Validation, Visualization
    \item Aayush Shah (\orcidlink{0000-0002-7029-2008} 0000-0002-7029-2008): Investigation, Validation
    \item Jenny Weng: (\orcidlink{0000-0002-1017-8908} 0000-0002-1017-8908): Investigation, Validation, Writing – review \& editing
    \item Rebekah Tang (\orcidlink{0000-0003-4563-5192} 0000-0003-4563-5192): Investigation
\end{itemize}

\bibliography{refs}

\begin{thebibliography}{}

\bibitem[Ali-Khan et~al., 2017]{ali_motivating_2017}
Ali-Khan, S.~E., Harris, L.~W., and Gold, E.~R. (2017).
\newblock Motivating participation in open science by examining researcher
  incentives.
\newblock {\em eLife}, 6, DOI:
  \href{https://dx.doi.org/10.7554/eLife.29319}{\ttfamily 10.7554/eLife.29319},
  \url{https://www.ncbi.nlm.nih.gov/pmc/articles/PMC5662284/}.

\bibitem[Amini et~al., 2019]{amini_uncovering_2019}
Amini, A., Soleimany, A.~P., Schwarting, W., Bhatia, S.~N., and Rus, D. (2019).
\newblock Uncovering and Mitigating Algorithmic Bias through Learned Latent
  Structure.
\newblock In {\em Proceedings of the 2019 {AAAI}/{ACM} Conference on {AI},
  Ethics, and Society}, {AIES} '19, pages 289--295. Association for Computing
  Machinery, DOI: \href{https://dx.doi.org/10.1145/3306618.3314243}{\ttfamily
  10.1145/3306618.3314243}, \url{https://doi.org/10.1145/3306618.3314243}.

\bibitem[Anderson et~al., 1976]{anderson1976land}
Anderson, J.~R., Hardy, E.~E., Roach, J.~T., and Witmer, R.~E. (1976).
\newblock {\em A Land Use and Land Cover Classification System for Use With
  Remote Sensor Data}, volume 964.
\newblock US Government Printing Office,
  \url{https://citeseerx.ist.psu.edu/viewdoc/download?doi=10.1.1.189.3029&rep=rep1&type=pdf}.

\bibitem[Baas et~al., 2020]{baas_scopus_2020}
Baas, J., Schotten, M., Plume, A., Côté, G., and Karimi, R. (2020).
\newblock Scopus as a curated, high-quality bibliometric data source for
  academic research in quantitative science studies.
\newblock {\em Quantitative Science Studies}, 1(1):377--386, DOI:
  \href{https://dx.doi.org/10.1162/qss_a_00019}{\ttfamily
  10.1162/qss\_a\_00019}, \url{https://doi.org/10.1162/qss_a_00019}.

\bibitem[Babbage, 1864]{babbage2011passages}
Babbage, C. (1864).
\newblock {\em Passages from the Life of a Philosopher}.
\newblock Longman, Green, Longman, Roberts, and Green, London.

\bibitem[Ball and Brunner, 2010]{ball_data_2010}
Ball, N.~M. and Brunner, R.~J. (2010).
\newblock Data Mining and Machine Learning in Astronomy.
\newblock {\em International Journal of Modern Physics D}, 19(7):1049--1106,
  DOI: \href{https://dx.doi.org/10.1142/S0218271810017160}{\ttfamily
  10.1142/S0218271810017160},
  \url{https://www.worldscientific.com/doi/abs/10.1142/S0218271810017160}.

\bibitem[Barclay et~al., 2019]{barclay2019towards}
Barclay, I., Preece, A., Taylor, I., and Verma, D. (2019).
\newblock Towards Traceability in Data Ecosystems using a Bill of Materials
  Model.
\newblock {\em arXiv preprint arXiv:1904.04253},
  \url{https://arxiv.org/abs/1904.04253}.

\bibitem[Bender and Friedman, 2018]{bender2018data}
Bender, E.~M. and Friedman, B. (2018).
\newblock Data statements for NLP: Toward mitigating system bias and enabling
  better science.
\newblock {\em Transactions of the ACL}, 6:587--604,
  \url{https://www.mitpressjournals.org/doi/pdf/10.1162/tacl_a_00041}.

\bibitem[Beretta et~al., 2018]{beretta2018ethical}
Beretta, E., Vetr{\`o}, A., Lepri, B., and De~Martin, J.~C. (2018).
\newblock Ethical and Socially-Aware Data Labels.
\newblock In {\em Annual International Symposium on Information Management and
  Big Data}, pages 320--327. Springer, DOI:
  \href{https://dx.doi.org/10.1007/978-3-030-11680-4_30}{\ttfamily
  10.1007/978-3-030-11680-4\_30}.

\bibitem[Bergstrom and West, 2020]{bergstrom_calling_2020}
Bergstrom, C.~T. and West, J.~D. (2020).
\newblock {\em Calling {Bullshit}: {The} {Art} of {Skepticism} in a
  {Data}-{Driven} {World}}.
\newblock Random House Publishing Group, London.

\bibitem[Birhane and Prabhu, 2021]{Birhane_2021_WACV}
Birhane, A. and Prabhu, V.~U. (2021).
\newblock Large Image Datasets: A Pyrrhic Win for Computer Vision?
\newblock In {\em Proceedings of the IEEE/CVF Winter Conference on Applications
  of Computer Vision (WACV)}, pages 1537--1547.
  \url{https://arxiv.org/abs/2006.16923}.

\bibitem[Blodgett et~al., 2020]{blodgett2020language}
Blodgett, S.~L., Barocas, S., Daum{\'e}~III, H., and Wallach, H. (2020).
\newblock Language (Technology) is Power: A Critical Survey of {``}Bias{''} in
  {NLP}.
\newblock In {\em Proceedings of the 58th Annual Meeting of the Association for
  Computational Linguistics}, pages 5454--5476, Online. Association for
  Computational Linguistics, DOI:
  \href{https://dx.doi.org/10.18653/v1/2020.acl-main.485}{\ttfamily
  10.18653/v1/2020.acl-main.485},
  \url{https://www.aclweb.org/anthology/2020.acl-main.485}.

\bibitem[Bontcheva et~al., 2013]{bontcheva_gate_2013}
Bontcheva, K., Cunningham, H., Roberts, I., Roberts, A., Tablan, V., Aswani,
  N., and Gorrell, G. (2013).
\newblock {GATE} {Teamware}: a web-based, collaborative text annotation
  framework.
\newblock {\em Language Resources and Evaluation}, 47(4):1007--1029, DOI:
  \href{https://dx.doi.org/10.1007/s10579-013-9215-6}{\ttfamily
  10.1007/s10579-013-9215-6}, \url{https://doi.org/10.1007/s10579-013-9215-6}.

\bibitem[Borgman, 2012]{borgman2012conundrum}
Borgman, C.~L. (2012).
\newblock The Conundrum of Sharing Research Data.
\newblock {\em Journal of the American Society for Information Science and
  Technology}, 63(6):1059--1078, \url{https://doi.org/10.1002/asi.22634}.

\bibitem[Bowker, 2005]{bowker_memory_2005}
Bowker, G. (2005).
\newblock {\em Memory Practices in the Sciences}.
\newblock The {MIT} Press, Cambridge, MA.

\bibitem[Bowker, 2020]{bowker_numbers_2020}
Bowker, G.~C. (2020).
\newblock Numbers or No Numbers in Science Studies.
\newblock {\em Quantitative Science Studies}, 1(3):927--929, DOI:
  \href{https://dx.doi.org/10.1162/qss_a_00054}{\ttfamily
  10.1162/qss\_a\_00054}, \url{https://doi.org/10.1162/qss_a_00054}.

\bibitem[Bowker and Star, 1999]{bowker1999sorting}
Bowker, G.~C. and Star, S.~L. (1999).
\newblock {\em Sorting Things Out: Classification and its Consequences}.
\newblock The MIT Press, Cambridge, MA.

\bibitem[Brady, 2016]{brady_error_2016}
Brady, A.~P. (2016).
\newblock Error and discrepancy in radiology: inevitable or avoidable?
\newblock {\em Insights into Imaging}, 8(1):171--182, DOI:
  \href{https://dx.doi.org/10.1007/s13244-016-0534-1}{\ttfamily
  10.1007/s13244-016-0534-1},
  \url{https://www.ncbi.nlm.nih.gov/pmc/articles/PMC5265198/}.

\bibitem[Brand et~al., 2015]{brand_beyond_2015}
Brand, A., Allen, L., Altman, M., Hlava, M., and Scott, J. (2015).
\newblock Beyond authorship: attribution, contribution, collaboration, and
  credit.
\newblock {\em Learned Publishing}, 28(2):151--155, DOI:
  \href{https://dx.doi.org/https://doi.org/10.1087/20150211}{\ttfamily
  https://doi.org/10.1087/20150211},
  \url{https://onlinelibrary.wiley.com/doi/abs/10.1087/20150211}.

\bibitem[Buolamwini and Gebru, 2018]{buolamwini2018gender}
Buolamwini, J. and Gebru, T. (2018).
\newblock Gender shades: Intersectional accuracy disparities in commercial
  gender classification.
\newblock In {\em ACM Conference on Fairness, Accountability and Transparency},
  pages 77--91. \url{http://proceedings.mlr.press/v81/buolamwini18a.html}.

\bibitem[Calmon et~al., 2017]{calmon_optimized_2017}
Calmon, F.~P., Wei, D., Vinzamuri, B., Ramamurthy, K.~N., and Varshney, K.~R.
  (2017).
\newblock Optimized Pre-Processing for Discrimination Prevention.
\newblock In {\em Proceedings of the 31st International Conference on Neural
  Information Processing Systems}, {NIPS}'17, pages 3995--4004. Curran
  Associates Inc.,
  \url{https://proceedings.neurips.cc/paper/2017/file/9a49a25d845a483fae4be7e341368e36-Paper.pdf}.

\bibitem[Cambrosio et~al., 2020]{cambrosio_beyond_2020}
Cambrosio, A., Cointet, J.-P., and Abdo, A.~H. (2020).
\newblock Beyond networks: Aligning qualitative and computational science
  studies.
\newblock {\em Quantitative Science Studies}, 1(3):1017--1024, DOI:
  \href{https://dx.doi.org/10.1162/qss_a_00055}{\ttfamily
  10.1162/qss\_a\_00055}, \url{https://doi.org/10.1162/qss_a_00055}.

\bibitem[Candela et~al., 2015]{candela_data_2015}
Candela, L., Castelli, D., Manghi, P., and Tani, A. (2015).
\newblock Data journals: A survey.
\newblock {\em Journal of the Association for Information Science and
  Technology}, 66(9):1747--1762, DOI:
  \href{https://dx.doi.org/https://doi.org/10.1002/asi.23358}{\ttfamily
  https://doi.org/10.1002/asi.23358},
  \url{https://asistdl.onlinelibrary.wiley.com/doi/abs/10.1002/asi.23358}.

\bibitem[Chang et~al., 2017]{chang_revolt_2017}
Chang, J.~C., Amershi, S., and Kamar, E. (2017).
\newblock Revolt: {Collaborative} {Crowdsourcing} for {Labeling} {Machine}
  {Learning} {Datasets}.
\newblock In {\em Proceedings of the 2017 {CHI} {Conference} on {Human}
  {Factors} in {Computing} {Systems}}, {CHI} '17, pages 2334--2346, New York,
  NY, USA. ACM, DOI:
  \href{https://dx.doi.org/10.1145/3025453.3026044}{\ttfamily
  10.1145/3025453.3026044}, \url{http://doi.acm.org/10.1145/3025453.3026044}.

\bibitem[Chavan and Penev, 2011]{chavan_data_2011}
Chavan, V. and Penev, L. (2011).
\newblock The data paper: a mechanism to incentivize data publishing in
  biodiversity science.
\newblock {\em {BMC} Bioinformatics}, 12:S2, DOI:
  \href{https://dx.doi.org/10.1186/1471-2105-12-S15-S2}{\ttfamily
  10.1186/1471-2105-12-S15-S2},
  \url{https://www.ncbi.nlm.nih.gov/pmc/articles/PMC3287445/}.

\bibitem[Costello, 2009]{costello_motivating_2009}
Costello, M.~J. (2009).
\newblock Motivating Online Publication of Data.
\newblock {\em {BioScience}}, 59(5):418--427, DOI:
  \href{https://dx.doi.org/10.1525/bio.2009.59.5.9}{\ttfamily
  10.1525/bio.2009.59.5.9},
  \url{https://bioone.org/journals/bioscience/volume-59/issue-5/bio.2009.59.5.9/Motivating-Online-Publication-of-Data/10.1525/bio.2009.59.5.9.full}.

\bibitem[Crawford and Paglen, 2019]{excavatingai}
Crawford, K. and Paglen, T. (2019).
\newblock Excavating AI: The Politics of Training Sets for Machine Learning.
\newblock \url{https://excavating.ai}.

\bibitem[Dastin, 2018]{dastin_amazon_2018}
Dastin, J. (2018).
\newblock Amazon scraps secret {AI} recruiting tool that showed bias against
  women.
\newblock {\em Reuters},
  \url{https://www.reuters.com/article/us-amazon-com-jobs-automation-insight-idUSKCN1MK08G}.

\bibitem[DeCamp and Lindvall, 2020]{decamp_latent_2020}
DeCamp, M. and Lindvall, C. (2020).
\newblock Latent bias and the implementation of artificial intelligence in
  medicine.
\newblock {\em Journal of the American Medical Informatics Association},
  27(12):2020--2023, DOI:
  \href{https://dx.doi.org/10.1093/jamia/ocaa094}{\ttfamily
  10.1093/jamia/ocaa094}, \url{https://doi.org/10.1093/jamia/ocaa094}.

\bibitem[Doddington et~al., 2004]{doddington2004automatic}
Doddington, G.~R., Mitchell, A., Przybocki, M.~A., Ramshaw, L.~A., Strassel,
  S.~M., and Weischedel, R.~M. (2004).
\newblock The Automatic Content Extraction (ACE) Program: Tasks, Data, and
  Evaluation.
\newblock In {\em Proceedings of the 2004 4th International Conference on
  Language Resources and Evaluation}, volume~2, pages 837--840, Paris. European
  Language Resources Association,
  \url{http://www.lrec-conf.org/proceedings/lrec2004/pdf/5.pdf}.

\bibitem[Dunn, 1961]{dunn1961multiple}
Dunn, O.~J. (1961).
\newblock Multiple Comparisons Among Means.
\newblock {\em Journal of the American Statistical Association},
  56(293):52--64.

\bibitem[Fecher and Friesike, 2014]{fecher_open_2014}
Fecher, B. and Friesike, S. (2014).
\newblock Open {Science}: {One} {Term}, {Five} {Schools} of {Thought}.
\newblock In Bartling, S. and Friesike, S., editors, {\em Opening {Science}:
  {The} {Evolving} {Guide} on {How} the {Internet} is {Changing} {Research},
  {Collaboration} and {Scholarly} {Publishing}}, pages 17--47. Springer
  International Publishing, Cham, DOI:
  \href{https://dx.doi.org/10.1007/978-3-319-00026-8_2}{\ttfamily
  10.1007/978-3-319-00026-8\_2},
  \url{https://doi.org/10.1007/978-3-319-00026-8_2}.

\bibitem[Fleming et~al., 2014]{fleming_blinded_2014}
Fleming, P.~S., Koletsi, D., and Pandis, N. (2014).
\newblock Blinded by {PRISMA}: Are Systematic Reviewers Focusing on {PRISMA}
  and Ignoring Other Guidelines?
\newblock {\em {PLOS} {ONE}}, 9(5):e96407, DOI:
  \href{https://dx.doi.org/10.1371/journal.pone.0096407}{\ttfamily
  10.1371/journal.pone.0096407},
  \url{https://journals.plos.org/plosone/article?id=10.1371/journal.pone.0096407}.

\bibitem[Fluke and Jacobs, 2020]{fluke_surveying_2020}
Fluke, C.~J. and Jacobs, C. (2020).
\newblock Surveying the reach and maturity of machine learning and artificial
  intelligence in astronomy.
\newblock {\em {WIREs} Data Mining and Knowledge Discovery}, 10(2):e1349, DOI:
  \href{https://dx.doi.org/https://doi.org/10.1002/widm.1349}{\ttfamily
  https://doi.org/10.1002/widm.1349},
  \url{https://onlinelibrary.wiley.com/doi/abs/10.1002/widm.1349}.

\bibitem[Friedler et~al., 2019]{friedler_comparative_2019}
Friedler, S.~A., Scheidegger, C., Venkatasubramanian, S., Choudhary, S.,
  Hamilton, E.~P., and Roth, D. (2019).
\newblock A Comparative Study of Fairness-Enhancing Interventions in Machine
  Learning.
\newblock In {\em Proceedings of the Conference on Fairness, Accountability,
  and Transparency}, {FAT}* '19, pages 329--338. Association for Computing
  Machinery, DOI: \href{https://dx.doi.org/10.1145/3287560.3287589}{\ttfamily
  10.1145/3287560.3287589}, \url{https://doi.org/10.1145/3287560.3287589}.

\bibitem[Friedman et~al., 2009]{friedman2009elements}
Friedman, J., Hastie, T., and Tibshirani, R. (2009).
\newblock {\em The Elements of Statistical Learning: Data Mining, Inference,
  and Prediction}.
\newblock Springer, New York, 2nd edition.

\bibitem[Gebru et~al., 2018]{gebru2018datasheets}
Gebru, T., Morgenstern, J., Vecchione, B., Vaughan, J.~W., Wallach, H.,
  Daume{\'e}~III, H., and Crawford, K. (2018).
\newblock Datasheets for Datasets.
\newblock {\em arXiv preprint arXiv:1803.09010},
  \url{https://arxiv.org/abs/1803.09010}.

\bibitem[Geiger et~al., 2020]{geiger2020garbage}
Geiger, R.~S., Yu, K., Yang, Y., Dai, M., Qiu, J., Tang, R., and Huang, J.
  (2020).
\newblock Garbage in, Garbage Out? Do Machine Learning Application Papers in
  Social Computing Report Where Human-labeled Training Data Comes From?
\newblock In {\em Proceedings of the 2020 Conference on Fairness,
  Accountability, and Transparency}, pages 325--336. DOI:
  \href{https://dx.doi.org/10.1145/3351095.3372862}{\ttfamily
  10.1145/3351095.3372862}, \url{https://arxiv.org/abs/1912.08320}.

\bibitem[Gewandter et~al., 2019]{gewandter_checklist_2019}
Gewandter, J.~S., Eisenach, J.~C., Gross, R.~A., Jensen, M.~P., Keefe, F.~J.,
  Lee, D.~A., and Turk, D.~C. (2019).
\newblock Checklist for the Preparation and Review of Pain Clinical Trial
  Publications: A Pain-Specific Supplement to {CONSORT}.
\newblock {\em Pain Reports}, 4(3):e621, DOI:
  \href{https://dx.doi.org/10.1097/PR9.0000000000000621}{\ttfamily
  10.1097/PR9.0000000000000621}.

\bibitem[Gharibi et~al., 2019]{gharibi_automated_2019}
Gharibi, G., Walunj, V., Alanazi, R., Rella, S., and Lee, Y. (2019).
\newblock Automated {Management} of {Deep} {Learning} {Experiments}.
\newblock In {\em Proceedings of the 3rd {International} {Workshop} on {Data}
  {Management} for {End}-to-{End} {Machine} {Learning}}, {DEEM}'19, pages
  8:1--8:4, New York, NY, USA. ACM, DOI:
  \href{https://dx.doi.org/10.1145/3329486.3329495}{\ttfamily
  10.1145/3329486.3329495}, \url{http://doi.acm.org/10.1145/3329486.3329495}.

\bibitem[Gil et~al., 2016]{gil_toward_2016}
Gil, Y., David, C.~H., Demir, I., Essawy, B.~T., Fulweiler, R.~W., Goodall,
  J.~L., Karlstrom, L., Lee, H., Mills, H.~J., Oh, J.-H., Pierce, S.~A., Pope,
  A., Tzeng, M.~W., Villamizar, S.~R., and Yu, X. (2016).
\newblock Toward the {Geoscience} {Paper} of the {Future}: {Best} practices for
  documenting and sharing research from data to software to provenance.
\newblock {\em Earth and Space Science}, 3(10):388--415, DOI:
  \href{https://dx.doi.org/10.1002/2015EA000136}{\ttfamily
  10.1002/2015EA000136},
  \url{https://agupubs.onlinelibrary.wiley.com/doi/abs/10.1002/2015EA000136}.

\bibitem[Gitelman, 2013]{gitelman2013raw}
Gitelman, L., editor (2013).
\newblock {\em Raw Data is an Oxymoron}.
\newblock The MIT Press, Cambridge, MA.

\bibitem[Goodfellow et~al., 2016]{goodfellow2016deep}
Goodfellow, I., Bengio, Y., and Courville, A. (2016).
\newblock {\em Deep Learning}.
\newblock The MIT Press, Cambridge, MA.
\newblock \url{http://www.deeplearningbook.org}.

\bibitem[Goodman et~al., 2014]{Goodman2014}
Goodman, A., Pepe, A., Blocker, A.~W., Borgman, C.~L., Cranmer, K., Crosas, M.,
  {Di Stefano}, R., Gil, Y., Groth, P., Hedstrom, M., Hogg, D.~W., Kashyap, V.,
  Mahabal, A., Siemiginowska, A., and Slavkovic, A. (2014).
\newblock {Ten Simple Rules for the Care and Feeding of Scientific Data}.
\newblock {\em PLoS Computational Biology}, 10(4):e1003542, DOI:
  \href{https://dx.doi.org/10.1371/journal.pcbi.1003542}{\ttfamily
  10.1371/journal.pcbi.1003542}.

\bibitem[Goodwin, 1994]{Goodwin1994}
Goodwin, C. (1994).
\newblock {Professional Vision}.
\newblock {\em American Anthropologist}, 96(3):606--633, DOI:
  \href{https://dx.doi.org/10.1525/aa.1994.96.3.02a00100}{\ttfamily
  10.1525/aa.1994.96.3.02a00100},
  \url{http://doi.wiley.com/10.1525/aa.1994.96.3.02a00100}.

\bibitem[Hanna et~al., 2020]{hanna_towards_2020}
Hanna, A., Denton, E., Smart, A., and Smith-Loud, J. (2020).
\newblock Towards a critical race methodology in algorithmic fairness.
\newblock In {\em Proceedings of the 2020 Conference on Fairness,
  Accountability, and Transparency}, {FAT}* '20, pages 501--512. Association
  for Computing Machinery, DOI:
  \href{https://dx.doi.org/10.1145/3351095.3372826}{\ttfamily
  10.1145/3351095.3372826}, \url{https://doi.org/10.1145/3351095.3372826}.

\bibitem[Hardt et~al., 2016]{hardt_equality_2016}
Hardt, M., Price, E., and Srebro, N. (2016).
\newblock Equality of opportunity in supervised learning.
\newblock In {\em Proceedings of the 30th International Conference on Neural
  Information Processing Systems}, {NIPS}'16, pages 3323--3331. Curran
  Associates Inc., DOI:
  \href{https://dx.doi.org/10.5555/3157382.3157469}{\ttfamily
  10.5555/3157382.3157469}.

\bibitem[Hind et~al., 2018]{hind2018increasing}
Hind, M., Mehta, S., Mojsilovic, A., Nair, R., Ramamurthy, K.~N., Olteanu, A.,
  and Varshney, K.~R. (2018).
\newblock Increasing Trust in AI Services through Supplier's Declarations of
  Conformity.
\newblock {\em arXiv preprint arXiv:1808.07261},
  \url{https://arxiv.org/pdf/1808.07261}.

\bibitem[Holland et~al., 2018]{holland2018dataset}
Holland, S., Hosny, A., Newman, S., Joseph, J., and Chmielinski, K. (2018).
\newblock The dataset nutrition label: A framework to drive higher data quality
  standards.
\newblock {\em arXiv preprint arXiv:1805.03677},
  \url{https://arxiv.org/abs/1805.03677}.

\bibitem[Hovy and Lavid, 2010]{hovy2010towards}
Hovy, E. and Lavid, J. (2010).
\newblock Towards a ‘science’ of corpus annotation: a new methodological
  challenge for corpus linguistics.
\newblock {\em International Journal of Translation}, 22(1):13--36.

\bibitem[Hunter, 2007]{Matplotlib}
Hunter, J.~D. (2007).
\newblock Matplotlib: A 2D Graphics Environment.
\newblock {\em Computing in Science \& Engineering}, 9(3):90--95, DOI:
  \href{https://dx.doi.org/10.1109/MCSE.2007.55}{\ttfamily
  10.1109/MCSE.2007.55},
  \url{https://aip.scitation.org/doi/abs/10.1109/MCSE.2007.55}.

\bibitem[Jacobs and Wallach, 2019]{jacobs_measurement_2019}
Jacobs, A.~Z. and Wallach, H. (2019).
\newblock Measurement and {Fairness}.
\newblock {\em arXiv:1912.05511 [cs]}, \url{http://arxiv.org/abs/1912.05511}.
\newblock arXiv: 1912.05511.

\bibitem[James et~al., 2013]{james2013introduction}
James, G., Witten, D., Hastie, T., and Tibshirani, R. (2013).
\newblock {\em An introduction to statistical learning}.
\newblock Springer, New York.

\bibitem[Jones et~al., 2001]{scipy}
Jones, E., Oliphant, T., Peterson, P., et~al. (2001).
\newblock {SciPy}: Open source scientific tools for {Python}.
\newblock \url{http://www.scipy.org/}.

\bibitem[Kang and Evans, 2020]{kang_against_2020}
Kang, D. and Evans, J. (2020).
\newblock Against method: Exploding the boundary between qualitative and
  quantitative studies of science.
\newblock {\em Quantitative Science Studies}, 1(3):930--944, DOI:
  \href{https://dx.doi.org/10.1162/qss_a_00056}{\ttfamily
  10.1162/qss\_a\_00056}, \url{https://doi.org/10.1162/qss_a_00056}.

\bibitem[Karimi~Mahabadi et~al., 2020]{karimi2020}
Karimi~Mahabadi, R., Belinkov, Y., and Henderson, J. (2020).
\newblock End-to-End Bias Mitigation by Modelling Biases in Corpora.
\newblock In {\em Proceedings of the 58th Annual Meeting of the Association for
  Computational Linguistics}, pages 8706--8716. Association for Computational
  Linguistics, DOI:
  \href{https://dx.doi.org/10.18653/v1/2020.acl-main.769}{\ttfamily
  10.18653/v1/2020.acl-main.769},
  \url{https://www.aclweb.org/anthology/2020.acl-main.769}.

\bibitem[Kitzes et~al., 2018]{Kitzes2018}
Kitzes, J., Turek, D., and Deniz, F. (2018).
\newblock {\em {The Practice of Reproducible Research : Case Studies and
  Lessons from the Data-Intensive Sciences}}.
\newblock University of California Press, Oakland,
  \url{http://practicereproducibleresearch.org}.

\bibitem[Kluyver et~al., 2016]{jupyter}
Kluyver, T., Ragan-Kelley, B., P{\'e}rez, F., Granger, B., Bussonnier, M.,
  Frederic, J., Kelley, K., Hamrick, J., Grout, J., Corlay, S., Ivanov, P.,
  Avila, D., Abdalla, S., and Willing, C. (2016).
\newblock Jupyter Notebooks: A Publishing format for Reproducible Computational
  Workflows.
\newblock In Loizides, F. and Schmidt, B., editors, {\em Positioning and Power
  in Academic Publishing: Players, Agents and Agendas}, pages 87 -- 90,
  Amsterdam. IOS Press, DOI:
  \href{https://dx.doi.org/10.3233/978-1-61499-649-1-87}{\ttfamily
  10.3233/978-1-61499-649-1-87}.

\bibitem[Krippendorff, 1970]{krippendorff1970}
Krippendorff, K. (1970).
\newblock Estimating the Reliability, Systematic Error and Random Error of
  Interval Data.
\newblock {\em Educational and Psychological Measurement}, 30(1):61--70, DOI:
  \href{https://dx.doi.org/10.1177/001316447003000105}{\ttfamily
  10.1177/001316447003000105}.

\bibitem[Krishnan et~al., 2016]{krishnan_activeclean_2016}
Krishnan, S., Franklin, M.~J., Goldberg, K., Wang, J., and Wu, E. (2016).
\newblock {ActiveClean}: {An} {Interactive} {Data} {Cleaning} {Framework} {For}
  {Modern} {Machine} {Learning}.
\newblock In {\em Proceedings of the 2016 {International} {Conference} on
  {Management} of {Data}}, {SIGMOD} '16, pages 2117--2120, New York, NY, USA.
  ACM, DOI: \href{https://dx.doi.org/10.1145/2882903.2899409}{\ttfamily
  10.1145/2882903.2899409}, \url{http://doi.acm.org/10.1145/2882903.2899409}.

\bibitem[Larivière et~al., 2020]{lariviere_investigating_2020}
Larivière, V., Pontille, D., and Sugimoto, C.~R. (2020).
\newblock Investigating the division of scientific labor using the Contributor
  Roles Taxonomy ({CRediT}).
\newblock {\em Quantitative Science Studies}, pages 1--18, DOI:
  \href{https://dx.doi.org/10.1162/qss_a_00097}{\ttfamily
  10.1162/qss\_a\_00097}, \url{https://doi.org/10.1162/qss_a_00097}.

\bibitem[Lary et~al., 2016]{lary_machine_2016}
Lary, D.~J., Alavi, A.~H., Gandomi, A.~H., and Walker, A.~L. (2016).
\newblock Machine learning in geosciences and remote sensing.
\newblock {\em Geoscience Frontiers}, 7(1):3--10, DOI:
  \href{https://dx.doi.org/10.1016/j.gsf.2015.07.003}{\ttfamily
  10.1016/j.gsf.2015.07.003},
  \url{https://www.sciencedirect.com/science/article/pii/S1674987115000821}.

\bibitem[Latour, 1987]{latour1987science}
Latour, B. (1987).
\newblock {\em Science in action: How to follow scientists and engineers
  through society}.
\newblock Harvard University Press, Cambridge, MA.

\bibitem[Leydesdorff et~al., 2020]{leydesdorff_bridging_2020}
Leydesdorff, L., Ràfols, I., and Milojević, S. (2020).
\newblock Bridging the divide between qualitative and quantitative science
  studies.
\newblock {\em Quantitative Science Studies}, 1(3):918--926, DOI:
  \href{https://dx.doi.org/10.1162/qss_e_00061}{\ttfamily
  10.1162/qss\_e\_00061}, \url{https://doi.org/10.1162/qss_e_00061}.

\bibitem[{Linguistic Data Consortium}, 2008]{linguistic2008ace}
{Linguistic Data Consortium} (2008).
\newblock ACE (Automatic Content Extraction) English annotation guidelines for
  entities version 6.6.
\newblock
  \url{https://www.ldc.upenn.edu/sites/www.ldc.upenn.edu/files/english-entities-guidelines-v6.6.pdf}.

\bibitem[Ma et~al., 2017]{ma_review_2017}
Ma, L., Li, M., Ma, X., Cheng, L., Du, P., and Liu, Y. (2017).
\newblock A review of supervised object-based land-cover image classification.
\newblock {\em {ISPRS} Journal of Photogrammetry and Remote Sensing},
  130:277--293, DOI:
  \href{https://dx.doi.org/10.1016/j.isprsjprs.2017.06.001}{\ttfamily
  10.1016/j.isprsjprs.2017.06.001},
  \url{https://www.sciencedirect.com/science/article/pii/S092427161630661X}.

\bibitem[McDonald et~al., 2019]{McDonald2019}
McDonald, N., Schoenebeck, S., and Forte, A. (2019).
\newblock Reliability and Inter-rater Reliability in Qualitative Research:
  Norms and Guidelines for CSCW and HCI Practice.
\newblock {\em Proc. ACM Hum.-Comput. Interact.}, 3(CSCW):72:1--72:23, DOI:
  \href{https://dx.doi.org/10.1145/3359174}{\ttfamily 10.1145/3359174},
  \url{http://doi.acm.org/10.1145/3359174}.

\bibitem[McKinney, 2010]{pandas}
McKinney, W. (2010).
\newblock {Data Structures for Statistical Computing in Python}.
\newblock In van~der Walt, S. and Millman, J., editors, {\em Proceedings of the
  9th Python in Science Conference}, pages 51--56.
  \url{http://conference.scipy.org/proceedings/scipy2010/mckinney.html}.

\bibitem[Medeiros and Ball, 2017]{Medeiros2017}
Medeiros, N. and Ball, R. (2017).
\newblock Teaching Integrity in Empirical Economics: The Pedagogy of
  Reproducible Science in Undergraduate Education.
\newblock In Hensley, M. and Davis-Kahl, S., editors, {\em Undergraduate
  Research and the Academic Librarian: Case Studies and Best Practices}.
  Association of College \& Research Libraries, Chicago,
  \url{https://scholarship.haverford.edu/cgi/viewcontent.cgi?article=1189}.

\bibitem[Mehrabi et~al., 2019]{mehrabi_survey_2019}
Mehrabi, N., Morstatter, F., Saxena, N., Lerman, K., and Galstyan, A. (2019).
\newblock A Survey on Bias and Fairness in Machine Learning.
\newblock \url{http://arxiv.org/abs/1908.09635}.

\bibitem[Mellin, 1957]{mellin1957work}
Mellin, W. (1957).
\newblock Work With New Electronic ‘Brains’ Opens Field for Army Math
  Experts.
\newblock {\em The Hammond Times}, 10:66.

\bibitem[Mitchell et~al., 2019]{mitchell2019model}
Mitchell, M., Wu, S., Zaldivar, A., Barnes, P., Vasserman, L., Hutchinson, B.,
  Spitzer, E., Raji, I.~D., and Gebru, T. (2019).
\newblock Model cards for model reporting.
\newblock In {\em Proceedings of the Conference on Fairness, Accountability,
  and Transparency}, pages 220--229. ACM, DOI:
  \href{https://dx.doi.org/10.1145/3287560.3287596}{\ttfamily
  10.1145/3287560.3287596}.

\bibitem[Mitchell, 1997]{mitchell1997}
Mitchell, T. (1997).
\newblock {\em Machine Learning}.
\newblock MacGraw-Hill, New York.

\bibitem[Moher et~al., 2009]{moher2009preferred}
Moher, D., Liberati, A., Tetzlaff, J., Altman, D.~G., and Group, P. (2009).
\newblock Preferred reporting items for systematic reviews and meta-analyses:
  the PRISMA statement.
\newblock {\em PLoS medicine}, 6(7):e1000097, DOI:
  \href{https://dx.doi.org/10.1371/journal.pmed.1000097}{\ttfamily
  10.1371/journal.pmed.1000097}.

\bibitem[Montgomery et~al., 2018]{montgomery_reporting_2018}
Montgomery, P., Grant, S., Mayo-Wilson, E., Macdonald, G., Michie, S.,
  Hopewell, S., Moher, D., Lawrence~Aber, J., Altman, D., Bhui, K., Booth, A.,
  Clark, D., Craig, P., Eisner, M., Fraser, M.~W., Gardner, F., Grant, S.,
  Hedges, L., Hollon, S., Hopewell, S., Kaplan, R., Kaufmann, P.,
  Konstantopoulos, S., Macdonald, G., Mayo-Wilson, E., McLeroy, K., Michie, S.,
  Mittman, B., Moher, D., Montgomery, P., Nezu, A., Sherman, L., Sonuga-Barke,
  E., Thomas, J., VandenBos, G., Waters, E., West, R., Yaffe, J., and {on
  behalf of the CONSORT-SPI Group} (2018).
\newblock Reporting Randomised Trials of Social and Psychological
  Interventions: the {CONSORT}-{SPI} 2018 Extension.
\newblock {\em Trials}, 19(1):407, DOI:
  \href{https://dx.doi.org/10.1186/s13063-018-2733-1}{\ttfamily
  10.1186/s13063-018-2733-1}, \url{https://doi.org/10.1186/s13063-018-2733-1}.

\bibitem[Mozeti{\v{c}} et~al., 2016]{Mozetic2016}
Mozeti{\v{c}}, I., Gr{\v{c}}ar, M., and Smailovi{\'{c}}, J. (2016).
\newblock {Multilingual Twitter Sentiment Classification: The Role of Human
  Annotators}.
\newblock {\em PLOS ONE}, 11(5):e0155036, DOI:
  \href{https://dx.doi.org/10.1371/journal.pone.0155036}{\ttfamily
  10.1371/journal.pone.0155036},
  \url{http://dx.plos.org/10.1371/journal.pone.0155036}.

\bibitem[Nakayama et~al., 2018]{doccano}
Nakayama, H., Kubo, T., Kamura, J., Taniguchi, Y., and Liang, X. (2018).
\newblock {doccano}: Text Annotation Tool for Human.
\newblock \url{https://github.com/doccano/doccano}.

\bibitem[Obermeyer et~al., 2019]{obermeyer_dissecting_2019}
Obermeyer, Z., Powers, B., Vogeli, C., and Mullainathan, S. (2019).
\newblock Dissecting racial bias in an algorithm used to manage the health of
  populations.
\newblock {\em Science}, 366(6464):447--453, DOI:
  \href{https://dx.doi.org/10.1126/science.aax2342}{\ttfamily
  10.1126/science.aax2342},
  \url{https://science.sciencemag.org/content/366/6464/447}.

\bibitem[Oleinik et~al., 2014]{oleinik_choice_2014}
Oleinik, A., Popova, I., Kirdina, S., and Shatalova, T. (2014).
\newblock On the choice of measures of reliability and validity in the
  content-analysis of texts.
\newblock {\em Quality \& Quantity}, 48(5):2703--2718, DOI:
  \href{https://dx.doi.org/10.1007/s11135-013-9919-0}{\ttfamily
  10.1007/s11135-013-9919-0}, \url{https://doi.org/10.1007/s11135-013-9919-0}.

\bibitem[Page and Moher, 2017]{page_evaluations_2017}
Page, M.~J. and Moher, D. (2017).
\newblock Evaluations of the Uptake and Impact of the Preferred Reporting Items
  for Systematic Reviews and Meta-Analyses ({PRISMA}) Statement and Extensions:
  A Scoping Review.
\newblock {\em Systematic Reviews}, 6(1):263, DOI:
  \href{https://dx.doi.org/10.1186/s13643-017-0663-8}{\ttfamily
  10.1186/s13643-017-0663-8}, \url{https://doi.org/10.1186/s13643-017-0663-8}.

\bibitem[Pandis et~al., 2015]{pandis_consort_2015}
Pandis, N., Fleming, P.~S., Hopewell, S., and Altman, D.~G. (2015).
\newblock The {CONSORT} Statement: Application Within and Adaptations for
  Orthodontic Trials.
\newblock {\em American Journal of Orthodontics and Dentofacial Orthopedics},
  147(6):663--679, DOI:
  \href{https://dx.doi.org/10.1016/j.ajodo.2015.03.014}{\ttfamily
  10.1016/j.ajodo.2015.03.014}.

\bibitem[P\'erez and Granger, 2007]{ipython}
P\'erez, F. and Granger, B.~E. (2007).
\newblock {IP}ython: a System for Interactive Scientific Computing.
\newblock {\em Computing in Science and Engineering}, 9(3):21--29, DOI:
  \href{https://dx.doi.org/10.1109/MCSE.2007.53}{\ttfamily
  10.1109/MCSE.2007.53}, \url{http://ipython.org}.

\bibitem[Perry, 2020]{simpledorff}
Perry, T. (2020).
\newblock SimpleDorff - Calculate Krippendorff's Alpha on a DataFrame.
\newblock \url{https://github.com/LightTag/simpledorff}.

\bibitem[Prabowo and Thelwall, 2009]{prabowo_sentiment_2009}
Prabowo, R. and Thelwall, M. (2009).
\newblock Sentiment analysis: A combined approach.
\newblock {\em Journal of Informetrics}, 3(2):143--157, DOI:
  \href{https://dx.doi.org/10.1016/j.joi.2009.01.003}{\ttfamily
  10.1016/j.joi.2009.01.003},
  \url{https://www.sciencedirect.com/science/article/pii/S1751157709000108}.

\bibitem[{P}roject {J}upyter et~al., 2018]{binder2018}
{P}roject {J}upyter, {M}atthias {B}ussonnier, {J}essica {F}orde, {J}eremy
  {F}reeman, {B}rian {G}ranger, {T}im {H}ead, {C}hris {H}oldgraf, {K}yle
  {K}elley, {G}ladys {N}alvarte, {A}ndrew {O}sheroff, {P}acer, M., {Y}uvi
  {P}anda, {F}ernando {P}erez, {B}enjamin~{R}agan {K}elley, and {C}arol
  {W}illing (2018).
\newblock {B}inder 2.0 - {R}eproducible, interactive, sharable environments for
  science at scale.
\newblock In {F}atih {A}kici, {D}avid {L}ippa, {D}illon {N}iederhut, and
  {P}acer, M., editors, {\em {P}roceedings of the 17th {P}ython in {S}cience
  {C}onference}, pages 113 -- 120. DOI:
  \href{https://dx.doi.org/10.25080/Majora-4af1f417-011}{\ttfamily
  10.25080/Majora-4af1f417-011}.

\bibitem[Pérez-Pérez et~al., 2015]{perez_marky_2015}
Pérez-Pérez, M., Glez-Peña, D., Fdez-Riverola, F., and Lourenço, A. (2015).
\newblock Marky: {A} tool supporting annotation consistency in multi-user and
  iterative document annotation projects.
\newblock {\em Computer Methods and Programs in Biomedicine}, 118(2):242--251,
  DOI: \href{https://dx.doi.org/10.1016/j.cmpb.2014.11.005}{\ttfamily
  10.1016/j.cmpb.2014.11.005},
  \url{http://www.sciencedirect.com/science/article/pii/S0169260714003848}.

\bibitem[Quarfoot and Levine, 2016]{quarfoot_how_2016}
Quarfoot, D. and Levine, R.~A. (2016).
\newblock How {Robust} {Are} {Multirater} {Interrater} {Reliability} {Indices}
  to {Changes} in {Frequency} {Distribution}?
\newblock {\em The American Statistician}, 70(4):373--384, DOI:
  \href{https://dx.doi.org/10.1080/00031305.2016.1141708}{\ttfamily
  10.1080/00031305.2016.1141708},
  \url{https://doi.org/10.1080/00031305.2016.1141708}.

\bibitem[Raji and Yang, 2019]{raji_about_2019}
Raji, I.~D. and Yang, J. (2019).
\newblock {ABOUT} {ML}: {Annotation} and {Benchmarking} on {Understanding} and
  {Transparency} of {Machine} {Learning} {Lifecycles}.
\newblock {\em arXiv:1912.06166 [cs, stat]},
  \url{http://arxiv.org/abs/1912.06166}.
\newblock arXiv: 1912.06166.

\bibitem[Ravi and Ravi, 2015]{ravi_survey_2015}
Ravi, K. and Ravi, V. (2015).
\newblock A survey on opinion mining and sentiment analysis: Tasks, approaches
  and applications.
\newblock {\em Knowledge-Based Systems}, 89:14--46, DOI:
  \href{https://dx.doi.org/10.1016/j.knosys.2015.06.015}{\ttfamily
  10.1016/j.knosys.2015.06.015},
  \url{https://www.sciencedirect.com/science/article/pii/S0950705115002336}.

\bibitem[Raykar and Yu, 2012]{raykar2012eliminating}
Raykar, V.~C. and Yu, S. (2012).
\newblock Eliminating spammers and ranking annotators for crowdsourced labeling
  tasks.
\newblock {\em Journal of Machine Learning Research}, 13(Feb):491--518, DOI:
  \href{https://dx.doi.org/10.5555/2188385.2188401}{\ttfamily
  10.5555/2188385.2188401}.

\bibitem[Rennie et~al., 2000]{rennie_contributions_2000}
Rennie, D., Flanagin, A., and Yank, V. (2000).
\newblock The Contributions of Authors.
\newblock {\em {JAMA}}, 284(1):89--91, DOI:
  \href{https://dx.doi.org/10.1001/jama.284.1.89}{\ttfamily
  10.1001/jama.284.1.89}, \url{https://doi.org/10.1001/jama.284.1.89}.

\bibitem[Riff et~al., 2013]{riff2013analyzing}
Riff, D., Lacy, S., and Fico, F. (2013).
\newblock {\em Analyzing media messages: Using quantitative content analysis in
  research}.
\newblock Routledge, New York.

\bibitem[Salimi et~al., 2020]{salimi_database_2020}
Salimi, B., Howe, B., and Suciu, D. (2020).
\newblock Database Repair Meets Algorithmic Fairness.
\newblock 49(1):34--41, DOI:
  \href{https://dx.doi.org/10.1145/3422648.3422657}{\ttfamily
  10.1145/3422648.3422657}, \url{https://doi.org/10.1145/3422648.3422657}.

\bibitem[Sallans and Donnelly, 2012]{sallans2012dmp}
Sallans, A. and Donnelly, M. (2012).
\newblock DMP Online and DMPTool: Different Strategies Towards a Shared Goal.
\newblock {\em International Journal of Digital Curation}, 7(2):123--129,
  \url{https://doi.org/10.2218/ijdc.v7i2.235}.

\bibitem[Sap et~al., 2019]{sap_risk_2019}
Sap, M., Card, D., Gabriel, S., Choi, Y., and Smith, N.~A. (2019).
\newblock The Risk of Racial Bias in Hate Speech Detection.
\newblock In {\em Proceedings of the 57th Annual Meeting of the Association for
  Computational Linguistics}, pages 1668--1678. Association for Computational
  Linguistics, DOI: \href{https://dx.doi.org/10.18653/v1/P19-1163}{\ttfamily
  10.18653/v1/P19-1163}, \url{https://www.aclweb.org/anthology/P19-1163}.

\bibitem[Schelter et~al., 2017]{schelter_automatically_2017}
Schelter, S., Böse, J.-H., Kirschnick, J., Klein, T., and Seufert, S. (2017).
\newblock Automatically Tracking Metadata and Provenance of Machine Learning
  Experiments.
\newblock In {\em Machine {Learning} {Systems} workshop at {NIPS}}.
  \url{http://learningsys.org/nips17/assets/papers/paper_13.pdf}.

\bibitem[Schelter et~al., 2018]{schelter_automating_2018}
Schelter, S., Lange, D., Schmidt, P., Celikel, M., Biessmann, F., and
  Grafberger, A. (2018).
\newblock Automating {Large}-scale {Data} {Quality} {Verification}.
\newblock {\em Proc. VLDB Endow.}, 11(12):1781--1794, DOI:
  \href{https://dx.doi.org/10.14778/3229863.3229867}{\ttfamily
  10.14778/3229863.3229867}, \url{https://doi.org/10.14778/3229863.3229867}.

\bibitem[Schreier et~al., 2006]{schreier2006academic}
Schreier, A.~A., Wilson, K., and Resnik, D. (2006).
\newblock Academic research record-keeping: Best practices for individuals,
  group leaders, and institutions.
\newblock {\em Academic medicine: journal of the Association of American
  Medical Colleges}, 81(1):42,
  \url{https://www.ncbi.nlm.nih.gov/pmc/articles/PMC3943904/}.

\bibitem[Schulz et~al., 2010]{schulz_consort_2010}
Schulz, K.~F., Altman, D.~G., Moher, D., and Group, f. t.~C. (2010).
\newblock {CONSORT} 2010 Statement: Updated Guidelines for Reporting Parallel
  Group Randomised Trials.
\newblock {\em {PLOS} Medicine}, 7(3):e1000251, DOI:
  \href{https://dx.doi.org/10.1371/journal.pmed.1000251}{\ttfamily
  10.1371/journal.pmed.1000251},
  \url{https://journals.plos.org/plosmedicine/article?id=10.1371/journal.pmed.1000251}.

\bibitem[Selbst et~al., 2019]{selbst_fairness_2019}
Selbst, A.~D., Boyd, D., Friedler, S.~A., Venkatasubramanian, S., and Vertesi,
  J. (2019).
\newblock Fairness and Abstraction in Sociotechnical Systems.
\newblock In {\em Proceedings of the Conference on Fairness, Accountability,
  and Transparency}, {FAT}* '19, pages 59--68. Association for Computing
  Machinery, DOI: \href{https://dx.doi.org/10.1145/3287560.3287598}{\ttfamily
  10.1145/3287560.3287598}, \url{https://doi.org/10.1145/3287560.3287598}.

\bibitem[Shipp et~al., 2002]{shipp_diffuse_2002}
Shipp, M.~A., Ross, K.~N., Tamayo, P., Weng, A.~P., Kutok, J.~L., Aguiar, R.
  C.~T., Gaasenbeek, M., Angelo, M., Reich, M., Pinkus, G.~S., Ray, T.~S.,
  Koval, M.~A., Last, K.~W., Norton, A., Lister, T.~A., Mesirov, J., Neuberg,
  D.~S., Lander, E.~S., Aster, J.~C., and Golub, T.~R. (2002).
\newblock Diffuse large B-cell lymphoma outcome prediction by gene-expression
  profiling and supervised machine learning.
\newblock {\em Nature Medicine}, 8(1):68--74, DOI:
  \href{https://dx.doi.org/10.1038/nm0102-68}{\ttfamily 10.1038/nm0102-68},
  \url{https://www.nature.com/articles/nm0102-68}.

\bibitem[Silberman et~al., 2018]{silberman2018responsible}
Silberman, M.~S., Tomlinson, B., LaPlante, R., Ross, J., Irani, L., and
  Zaldivar, A. (2018).
\newblock Responsible Research With Crowds: Pay Crowdworkers at Least Minimum
  Wage.
\newblock {\em Communications of the ACM}, 61(3):39--41, DOI:
  \href{https://dx.doi.org/10.1145/3180492}{\ttfamily 10.1145/3180492}.

\bibitem[Simpson et~al., 2014]{Simpson2014}
Simpson, R., Page, K.~R., and De~Roure, D. (2014).
\newblock Zooniverse: Observing the World's Largest Citizen Science Platform.
\newblock In {\em Proceedings of the 23rd International Conference on World
  Wide Web}, WWW '14 Companion, pages 1049--1054, New York, NY, USA. ACM, DOI:
  \href{https://dx.doi.org/10.1145/2567948.2579215}{\ttfamily
  10.1145/2567948.2579215}, \url{http://doi.acm.org/10.1145/2567948.2579215}.

\bibitem[Singh et~al., 2019]{singh_decision_2019}
Singh, J., Cobbe, J., and Norval, C. (2019).
\newblock Decision {Provenance}: {Harnessing} {Data} {Flow} for {Accountable}
  {Systems}.
\newblock {\em IEEE Access}, 7:6562--6574, DOI:
  \href{https://dx.doi.org/10.1109/ACCESS.2018.2887201}{\ttfamily
  10.1109/ACCESS.2018.2887201}.

\bibitem[Skitka et~al., 1999]{skitka1999does}
Skitka, L.~J., Mosier, K.~L., and Burdick, M. (1999).
\newblock Does automation bias decision-making?
\newblock {\em International Journal of Human-Computer Studies},
  51(5):991--1006, DOI:
  \href{https://dx.doi.org/10.1006/ijhc.1999.0252}{\ttfamily
  10.1006/ijhc.1999.0252}.

\bibitem[Smaldino, 2016]{smaldino_why_2016}
Smaldino, P. (2016).
\newblock Why isn’t science better? Look at career incentives.
\newblock {\em The Conversation},
  \url{http://theconversation.com/why-isnt-science-better-look-at-career-incentives-65619}.

\bibitem[Smith, 2009]{smith_data_2009}
Smith, V.~S. (2009).
\newblock Data publication: towards a database of everything.
\newblock {\em {BMC} Research Notes}, 2:113, DOI:
  \href{https://dx.doi.org/10.1186/1756-0500-2-113}{\ttfamily
  10.1186/1756-0500-2-113},
  \url{https://www.ncbi.nlm.nih.gov/pmc/articles/PMC2702265/}.

\bibitem[Sober{\'o}n et~al., 2013]{soberon2013measuring}
Sober{\'o}n, G., Aroyo, L., Welty, C., Inel, O., Lin, H., and Overmeen, M.
  (2013).
\newblock Measuring crowd truth: Disagreement metrics combined with worker
  behavior filters.
\newblock In {\em CrowdSem 2013 Workshop}.
  \url{http://ceur-ws.org/Vol-1030/paper-07.pdf}.

\bibitem[Thelwall et~al., 2010]{thelwall_sentiment_2010}
Thelwall, M., Buckley, K., Paltoglou, G., Cai, D., and Kappas, A. (2010).
\newblock Sentiment strength detection in short informal text.
\newblock {\em Journal of the American Society for Information Science and
  Technology}, 61(12):2544--2558, DOI:
  \href{https://dx.doi.org/https://doi.org/10.1002/asi.21416}{\ttfamily
  https://doi.org/10.1002/asi.21416},
  \url{https://asistdl.onlinelibrary.wiley.com/doi/abs/10.1002/asi.21416}.

\bibitem[Tinsley and Weiss, 1975]{tinsley1975interrater}
Tinsley, H.~E. and Weiss, D.~J. (1975).
\newblock Interrater reliability and agreement of subjective judgments.
\newblock {\em Journal of Counseling Psychology}, 22(4):358, DOI:
  \href{https://dx.doi.org/10.1037/h0076640}{\ttfamily 10.1037/h0076640}.

\bibitem[Tong et~al., 2007]{Tong2007}
Tong, A., Sainsbury, P., and Craig, J. (2007).
\newblock {Consolidated criteria for reporting qualitative research (COREQ): a
  32-item checklist for interviews and focus groups}.
\newblock {\em International Journal for Quality in Health Care},
  19(6):349--357, DOI:
  \href{https://dx.doi.org/10.1093/intqhc/mzm042}{\ttfamily
  10.1093/intqhc/mzm042},
  \url{https://academic.oup.com/intqhc/article-lookup/doi/10.1093/intqhc/mzm042}.

\bibitem[van~der Walt et~al., 2011]{numpy}
van~der Walt, S., Colbert, S.~C., and Varoquaux, G. (2011).
\newblock The NumPy Array: A Structure for Efficient Numerical Computation.
\newblock {\em Computing in Science Engineering}, 13(2):22--30, DOI:
  \href{https://dx.doi.org/10.1109/MCSE.2011.37}{\ttfamily
  10.1109/MCSE.2011.37}, \url{https://arxiv.org/abs/1102.1523}.

\bibitem[van Rossum, 1995]{python}
van Rossum, G. (1995).
\newblock Python Library Reference.
\newblock \url{https://ir.cwi.nl/pub/5009/05009D.pdf}.

\bibitem[Vayena et~al., 2018]{vayena_machine_2018}
Vayena, E., Blasimme, A., and Cohen, I.~G. (2018).
\newblock Machine learning in medicine: {Addressing} ethical challenges.
\newblock {\em PLOS Medicine}, 15(11):e1002689, DOI:
  \href{https://dx.doi.org/10.1371/journal.pmed.1002689}{\ttfamily
  10.1371/journal.pmed.1002689},
  \url{https://journals.plos.org/plosmedicine/article?id=10.1371/journal.pmed.1002689}.

\bibitem[Waskom et~al., 2018]{seaborn}
Waskom, M., Botvinnik, O., O'Kane, D., Hobson, P., Ostblom, J., Lukauskas, S.,
  Gemperline, D.~C., Augspurger, T., Halchenko, Y., Cole, J.~B., Warmenhoven,
  J., de~Ruiter, J., Pye, C., Hoyer, S., Vanderplas, J., Villalba, S., Kunter,
  G., Quintero, E., Bachant, P., Martin, M., Meyer, K., Miles, A., Ram, Y.,
  Brunner, T., Yarkoni, T., Williams, M.~L., Evans, C., Fitzgerald, C., Brian,
  and Qalieh, A. (2018).
\newblock Seaborn: Statistical Data Visualization Using Matplotlib.
\newblock DOI: \href{https://dx.doi.org/10.5281/zenodo.592845}{\ttfamily
  10.5281/zenodo.592845}, \url{https://seaborn.pydata.org}.

\bibitem[Welch, 1947]{welch1947generalization}
Welch, B.~L. (1947).
\newblock The Generalization of Student's Problem When Several Different
  Population Variances Are Involved.
\newblock {\em Biometrika}, 34(1/2):28--35, DOI:
  \href{https://dx.doi.org/10.2307/2332510}{\ttfamily 10.2307/2332510}.

\bibitem[Wilson et~al., 2017]{Wilson2017}
Wilson, G., Bryan, J., Cranston, K., Kitzes, J., Nederbragt, L., and Teal,
  T.~K. (2017).
\newblock {Good enough practices in scientific computing}.
\newblock {\em PLOS Computational Biology}, 13(6):e1005510, DOI:
  \href{https://dx.doi.org/10.1371/journal.pcbi.1005510}{\ttfamily
  10.1371/journal.pcbi.1005510}.

\bibitem[Wu and Zhang, 2016]{wu_automated_2016}
Wu, X. and Zhang, X. (2016).
\newblock Automated {Inference} on {Criminality} using {Face} {Images}.
\newblock {\em arXiv:1611.04135 [cs]}, \url{http://arxiv.org/abs/1611.04135}.
\newblock arXiv: 1611.04135, version: 3.

\bibitem[Ye et~al., 2003]{ye_predicting_2003}
Ye, Q.-H., Qin, L.-X., Forgues, M., He, P., Kim, J.~W., Peng, A.~C., Simon, R.,
  Li, Y., Robles, A.~I., Chen, Y., Ma, Z.-C., Wu, Z.-Q., Ye, S.-L., Liu, Y.-K.,
  Tang, Z.-Y., and Wang, X.~W. (2003).
\newblock Predicting hepatitis B virus–positive metastatic hepatocellular
  carcinomas using gene expression profiling and supervised machine learning.
\newblock {\em Nature Medicine}, 9(4):416--423, DOI:
  \href{https://dx.doi.org/10.1038/nm843}{\ttfamily 10.1038/nm843},
  \url{https://www.nature.com/articles/nm843}.

\bibitem[Zafar et~al., 2017]{zafar_fairness_2017}
Zafar, M.~B., Valera, I., Rogriguez, M.~G., and Gummadi, K.~P. (2017).
\newblock Fairness Constraints: Mechanisms for Fair Classification.
\newblock In {\em Artificial Intelligence and Statistics}, pages 962--970.
  {PMLR}, \url{http://proceedings.mlr.press/v54/zafar17a.html}.
\newblock {ISSN}: 2640-3498.

\bibitem[Zimring, 2019]{zimring_were_2019}
Zimring, J. (2019).
\newblock We're Incentivizing Bad Science.
\newblock {\em Scientific American},
  \url{https://blogs.scientificamerican.com/observations/were-incentivizing-bad-science/}.

\bibitem[Zuckerman, 2020]{zuckerman_is_2020}
Zuckerman, H. (2020).
\newblock Is “the time ripe” for quantitative research on misconduct in
  science?
\newblock {\em Quantitative Science Studies}, 1(3):945--958, DOI:
  \href{https://dx.doi.org/10.1162/qss_a_00065}{\ttfamily
  10.1162/qss\_a\_00065}, \url{https://doi.org/10.1162/qss_a_00065}.

\end{thebibliography}
\appendix
\clearpage
\section{Corpus details}
The tables below describe each corpus according to metadata provided by Scopus.
\begin{table}[h!]
\caption{Publication year of papers by corpus}
\label{tab:date}
\begin{tabular}{|l|l|l|l|l|}
\hline
\textbf{Year} & \textbf{Biomedical} & \textbf{Physical \& Enviro Sci} & \textbf{Social Science} & \textbf{Grand Total} \\ \hline
\textbf{2013} & 2 & 3 & 4 & 9 \\ \hline
\textbf{2014} & 6 & 4 & 6 & 16 \\ \hline
\textbf{2015} & 6 & 7 & 5 & 18 \\ \hline
\textbf{2016} & 18 & 13 & 10 & 41 \\ \hline
\textbf{2017} & 14 & 15 & 13 & 42 \\ \hline
\textbf{2018} & 14 & 28 & 32 & 74 \\ \hline
\textbf{Grand Total} & \textbf{60} & \textbf{70} & \textbf{70} & \textbf{200} \\ \hline
\end{tabular}
\vspace{1cm}
\caption{Classifier area/domain by corpus}
\label{tab:outcome-corpus}
\begin{tabular}{|l|l|l|l|l|}
\hline
\textbf{} & \textbf{Biomedical} & \textbf{Physical \& Enviro Sci} & \textbf{Social Science} & Grand Total \\ \hline
\textbf{Activities \& actions} & 2 & 2 & 3 & 7 \\ \hline
\textbf{Biological} & 10 & 6 & 1 & 17 \\ \hline
\textbf{Demographic} & 0 & 2 & 3 & 5 \\ \hline
\textbf{Geo/ecological} & 1 & 9 & 3 & 13 \\ \hline
\textbf{Linguistic} & 2 & 3 & 19 & 24 \\ \hline
\textbf{Medical} & 25 & 12 & 6 & 43 \\ \hline
\textbf{Other} & 0 & 2 & 2 & 4 \\ \hline
\textbf{Physical} & 4 & 8 & 2 & 14 \\ \hline
\textbf{Soft/hardware} & 1 & 8 & 5 & 14 \\ \hline
\textbf{N/A} & 15 & 18 & 26 & 59 \\ \hline
\textbf{Grand Total} & 60 & 70 & 70 & 200 \\ \hline
\end{tabular}
\vspace{1cm}

\caption{Publication type by corpus (note: all conference papers were peer reviewed according to Scopus)}
\label{tab:pub-type-corpus}
\begin{tabular}{|l|l|l|l|l|}
\hline
 & Biomedical & Physical \& Enviro Sci & Social Science & Grand Total \\ \hline
Journal Article & 50 & 40 & 31 & 121 \\ \hline
Conference Paper & 10 & 30 & 39 & 79 \\ \hline
Grand Total & 60 & 70 & 70 & 200 \\ \hline
\end{tabular}
\end{table}

\section{Labeling protocol and instructions}
These are the instructions that were given to labelers. Note that we have removed links to specific papers that were used as examples, which can be provided upon request.

{\parindent0pt 
{\setlength{\parskip}{1em}
\subsection{General purpose info}\label{general-purpose-info}

Put N/A if the question is not applicable. Some questions you put a -
for no information.

``Unsure'' is always an acceptable answer. It is OK to say that you are
unsure. Some things are vague or complicated. Some of these papers are
bad, wrong, messy, use words in completely different ways than is
standard, shouldn't have been published at all, aren't actually about
machine learning, and so on.

Flag complicated cases for discussion, use the last notes column. Some
papers might have weird cases that require us to redefine our
instructions, which is also OK. Don't spend too much time agonizing over
small decisions.

``Coding'' can mean manual annotation, this is a legacy from early 20th
century linguistics.

If they have a link or citation to more info about their annotation,
follow that. If it is an existing dataset from another project, that is
out of scope. Don't try to search deeply for it if it isn't referenced
in the paper.

If the paper is over 50 pages, skip it and flag for discussion.

If two rounds/stages of human annotation are involved, use the first
one.

If there are multiple independent classifiers, some using human labels
and others using machine labels, focus on the human-labeled classifier
and ignore the machine-labeled one.

\subsection{Questions}\label{questions}

\subsubsection{Original ML classification task:} Is the paper presenting its
own original classifier that is trying to predict something?
``Original'' means a new classifier they made based on new or old data,
not anything about the novelty or innovation in the problem area.

\begin{itemize}
\item
  \begin{quote}
  Yes
  \end{quote}
\item
  \begin{quote}
  No
  \end{quote}
\item
  \begin{quote}
  Unsure / no information
  \end{quote}
\end{itemize}

Classification involves predicting cases on a defined set of categories.
Prediction is required, but not enough. Linear regressions might be
included if the regression is used to make a classification, but making
predictions for a linear variable is not. Predicting income or age
brackets is classification, predicting raw income or age is not.

Example: if the paper is vague about if they actually built a
classifier, choose unsure (link removed)

Example: any prediction on a linear/scalar value (not binned categories)
is not classification (link removed)

Example: recommender systems are typically not classification (link
removed)

Example: analyzing statistics about the kinds of words people use on
social media is not a classification task

Example: predicting location is a classification task if it is from a
set of locations (from work, school, home, or other) but not if it is an
infinite/undefined number of locations. (link removed)

Example: This paper (link removed) was framed as not an original
classification task (more algorithm performance), but they did create an
original classifier. This can also be an ``unsure'' -\/- which is 100\%
OK to answer.

Example: Literature review papers that include classification papers
aren't in this, if they didn't actually build a classifier.

Example: if there is a supervised classification task that is part of a
broader process, this counts, focus on that.

If no, skip the following questions.

\subsubsection{Classification outcome:} What is the general type of problem or
outcome that the classifier is trying to predict? This will be the
label, typically. What is the end result, not how did they get there.

\textbf{If multiple apply, put both and separate by comma}

\begin{itemize}
\item
  \begin{quote}
  Linguistic outcome
  \end{quote}

  \begin{itemize}
  \item
    \begin{quote}
    Sentiment, part of speech, stance, sarcasm, language, spam (if using
    NLP)
    \end{quote}
  \end{itemize}
\item
  \begin{quote}
  Medical diagnosis
  \end{quote}
\item
  \begin{quote}
  Biological classification (non-medical)
  \end{quote}
\item
  \begin{quote}
  Physical classification
  \end{quote}

  \begin{itemize}
  \item
    \begin{quote}
    Astronomy, light, materials, chemicals, voltage
    \end{quote}
  \end{itemize}
\item
  \begin{quote}
  Demographic classification
  \end{quote}

  \begin{itemize}
  \item
    \begin{quote}
    gender, race, class, political, job, transport
    \end{quote}
  \end{itemize}
\item
  \begin{quote}
  Ecological/geographic/land use
  \end{quote}

  \begin{itemize}
  \item
    \begin{quote}
    Buildings
    \end{quote}
  \end{itemize}
\item
  \begin{quote}
  Activities and actions
  \end{quote}

  \begin{itemize}
  \item
    \begin{quote}
    What are they doing? Includes gesture identification, tweeting while
    drinking, fraud/price manipulation
    \end{quote}
  \end{itemize}
\item
  \begin{quote}
  Software and hardware entities
  \end{quote}

  \begin{itemize}
  \item
    \begin{quote}
    Includes various technologies that don't fit the above categories
    (esp not activities / actions)
    \end{quote}
  \item
    \begin{quote}
    Bot and other malware detection
    \end{quote}
  \end{itemize}
\item
  \begin{quote}
  Other
  \end{quote}
\end{itemize}

\subsubsection{Labels from human annotation:} Is the classifier at least in
part trained on labeled data that involves human(s) who make individual
judgments for specific items? This requires a human to make a judgement
about individual cases.

\begin{itemize}
\item
  \begin{quote}
  \textbf{Yes for all}
  \end{quote}

  \begin{itemize}
  \item
    \begin{quote}
    This is the typical case, where every item in the dataset used to
    train the classifier had a human make a judgment about the item
    \end{quote}
  \end{itemize}
\item
  \begin{quote}
  \textbf{Yes for some}, which applies if *either*:
  \end{quote}

  \begin{itemize}
  \item
    \begin{quote}
    Human annotation was used to evaluate the classifier, but not train
    it; or,
    \end{quote}
  \item
    \begin{quote}
    There was some process of humans making judgments about some items,
    but then some kind of automated / mechanical way of scaling this to
    all items.
    \end{quote}
  \end{itemize}
\item
  \begin{quote}
  \textbf{No / machine-labeled}
  \end{quote}

  \begin{itemize}
  \item
    \begin{quote}
    This includes fully-automated / machine-labeled ways of extracting
    labels, where a human is not involved in making each individual
    judgment
    \end{quote}
  \end{itemize}
\item
  \begin{quote}
  \textbf{Implicit yes}
  \end{quote}

  \begin{itemize}
  \item
    \begin{quote}
    We know based on the subject matter that it had to be human labeled
    (e.g. patient medical data: 10.1109/CITSM.2017.8089245)
    \end{quote}
  \end{itemize}
\item
  \begin{quote}
  \textbf{No information}
  \end{quote}

  \begin{itemize}
  \item
    \begin{quote}
    When we know for sure that there is no info in the paper and the
    context doesn't necessarily imply human labels (e.g.
    https://doi.org/10.1016/j.jtbi.2016.05.011)
    \end{quote}
  \end{itemize}
\item
  \begin{quote}
  \textbf{Unsure}
  \end{quote}

  \begin{itemize}
  \item
    \begin{quote}
    This is when we are so confused by the paper that we don't even know
    if we can or can't answer it
    \end{quote}
  \end{itemize}
\end{itemize}

If a human is told to do something that is recorded, where the
recordings are the data and what they were told to do is the label, this
counts as labels from human annotation. (e.g. (link removed) and (link
removed))

This includes re-using existing data from human judgements, if it was
for the same purpose as the classifier. This does not include clever
re-using of metadata or human labeled data for a new purpose.

Setting a threshold for quantitative data is not human-labeled.

Do a quick CTRL-F for ``manual'', ``annot'', and ``label'' if you don't
see anything, just to be sure.

Example (link removed): this is yes for some, but very implicit.
Unsupervised clustering used to re-label

Example: In medicine, if they were classifying for high blood pressure,
and they use existing patient records with a cutoff of some number, this
is not human annotation, it is ``no / machine labeled''. If they extract
a physician's judgment from medical records, it is.

Example: labels were length of stay, not human annotation: (link
removed)

\begin{quote}
BUT: psych diagnostic tests that require judgement in scoring beyond
skill are human judgement, although this is borderline.
\end{quote}

Example: If the paper is using an external dataset that we know
implicitly would have to use human annotation, but they don't say
anything, put ``Implicit yes.'' (link removed)

Example: paper on political stances was labels from human annotation,
just not original. They took the labels from elsewhere and filled in the
gaps (more on that in next Q).

Example: Buying followers and seeing who follows (link removed) is not
human annotation.

Example: Generating (smart) simulated datasets from metadata is not
human annotation.

Example: (link removed) not annotation when looking up political
affiliation of politicians, even though it is manual work. No judgement
is involved. ``No.''

Example: (link removed) identified hashtags that they believe
universally correspond to certain political stances. This would be a
kind of ``self-annotation'' by the tweet's author and therefore ``yes
for all''

Example: If they are using human annotation to have confidence that a
machine-annotated dataset is as good as a human annotated one, but the
human annotated dataset isn't actually used to train the classifier, it
counts as labels from human annotation -\/-\/- see next question

Example: (link removed) did an inductive approach to finding accounts
they thought were hate accounts, then used those tweets for training
data. This is classification and human annotation.

Example: (link removed) -\/- they recruited patients from a parkinsons'
clinic for evaluation, their classifier doesn't say what it was trained
on. Put as ``Yes, for some items''

On multi-stage processes: in general, focus on where humans are most in
the loop with labeling. If there are multiple independent classifiers,
some using human labels and others using machine labels, focus on the
human-labeled classifier and ignore the machine-labeled one. E.g. (link
removed) should be ``yes for all'' because we're just looking at the
human-labeled one.

Unsure doesn't necessarily mean low quality: (link removed) -\/- eg
cases where human judgement can be done at scale / semi-automated /
rules

If not, skip the following questions about human annotation.

\subsubsection{Used human annotation for training vs. evaluation:} If human
annotation was used to generate labels, was it for the training dataset
or just for evaluation?

\begin{itemize}
\item
  \begin{quote}
  \textbf{No human annotation}
  \end{quote}
\item
  \begin{quote}
  \textbf{Human annotation for training data only}
  \end{quote}

  \begin{itemize}
  \item
    \begin{quote}
    This is the typical case, where labels are created then used to
    train the classifier. Often part of this data is held out as a test
    set to evaluate the classifier
    \end{quote}
  \end{itemize}
\item
  \begin{quote}
  \textbf{Human annotation for evaluation only}
  \end{quote}

  \begin{itemize}
  \item
    \begin{quote}
    This is when they train the classifier using non human-labeled data,
    but use humans to either evaluate the validity of that dataset or
    the classifier.
    \end{quote}
  \end{itemize}
\item
  \begin{quote}
  \textbf{Unsure}
  \end{quote}
\end{itemize}

\subsubsection{Used original human annotation:} Did the project involve
creating new human-labeled data (original), or was it exclusively
re-using an existing dataset (external), or both? Think about this about
as being organized by the researcher.

\begin{itemize}
\item
  \begin{quote}
  \textbf{No original human annotation}
  \end{quote}
\item
  \begin{quote}
  \textbf{Only used original human annotation}
  \end{quote}
\item
  \begin{quote}
  \textbf{Only used external human annotation}
  \end{quote}
\item
  \begin{quote}
  \textbf{Used both original and external human annotation}
  \end{quote}
\item
  \begin{quote}
  \textbf{Unsure}
  \end{quote}
\end{itemize}

Papers may have a mix of new and old human labeled data, or new human
labeled data and non-human labeled data.

New human annotation must be systematic, not filling in the gaps of
another dataset. Example: [removed] paper on political stances is *not*
original human annotation, even though they did some manual original
research to fill the gap.

For surveys, this counts as original if they ran the survey themselves,
external if re-using someone else's survey data.

If the methods section is too vague to not tell, then leave as unsure
(link removed)

If using transfer learning to augment a human-labeled classifier, this
is original and external.

There may be overlap between the authors of a paper and a previous paper
that presented the dataset. If a paper is treating a dataset/paper as a
separate paper, consider it external, regardless of author overlap, even
if they use language like ``we previously.'' Example (link removed)

\textbf{ONLY CONTINUE IF SOURCE INCLUDES ORIGINAL HUMAN ANNOTATION}

\textbf{Put N/A after for all items if not}

\subsubsection{Original human annotation source:} Who were the human
annotators? Drop-down options are:

\begin{itemize}
\item
  \begin{quote}
  \textbf{Amazon Mechanical Turk (AMT, Turkers)}
  \end{quote}
\item
  \begin{quote}
  \textbf{Other crowdworking platform (Crowdflower / Figure8)}
  \end{quote}
\item
  \begin{quote}
  \textbf{The paper's authors}
  \end{quote}
\item
  \begin{quote}
  \textbf{Other w/ claim of expertise}
  \end{quote}
\item
  \begin{quote}
  \textbf{Students (w/ no claim of expertise)}
  \end{quote}
\item
  \begin{quote}
  \textbf{Other (w/ no claim of expertise)}
  \end{quote}
\item
  \begin{quote}
  \textbf{Survey / self-reported data}
  \end{quote}
\item
  \begin{quote}
  \textbf{No information in the paper}
  \end{quote}
\item
  \begin{quote}
  \textbf{Unsure}
  \end{quote}
\end{itemize}

Survey / self reported includes self-annotation from social media (e.g.
emoticons, hashtags), unless it was unstructured data that was then
labeled by others. This also includes if a human is told to do something
that is recorded, where the recordings are the data and what they were
told to do is the label, this counts as labels from human annotation.
(e.g. (link removed)

Do not consider data cleaning as part of annotation, unless it was
systematic and involved looking at all or a random sample of the data.

``Other w/ claim of expertise'' involves a specific claim that the
annotators had qualifications beyond the average person. This is
independent from the kinds of specific training they received for the
task at hand.

Example: (link removed) psychiatrists made a determination of patients
having or not having an eating disorder, based on surveys of patients.
This is expert, not survey. But if they just had patients take one
survey and the authors put them in different labels, this would be
survey / self-reported.

Example: (link removed) they had volunteers collect data about road
quality from smartphones, then they annotated their own data with
labels. This is ``survey / self-reported''

\subsubsection{Prescreening for crowdwork platforms }

Put N/A if this is not applicable, if they are not using crowdwork.

\begin{itemize}
\item
  \begin{quote}
  \textbf{No prescreening (must state this)}
  \end{quote}
\item
  \begin{quote}
  \textbf{Previous platform performance qualification (e.g. AMT Master)}
  \end{quote}
\item
  \begin{quote}
  \textbf{Generic skills-based qualification (e.g. AMT Premium)}
  \end{quote}
\item
  \begin{quote}
  \textbf{Location qualification (country)}
  \end{quote}
\item
  \begin{quote}
  \textbf{Project-specific prescreening (e.g. inviting good crowdworkers
  back, doing their own prescreening)}
  \end{quote}
\item
  \begin{quote}
  \textbf{No information}
  \end{quote}
\item
  \begin{quote}
  \textbf{Unsure}
  \end{quote}
\end{itemize}

\subsubsection{Annotator compensation:} Does the paper discuss how the
annotators were compensated, if at all? If more than one applies, enter
manually: list all and separate by commas

\begin{itemize}
\item
  \begin{quote}
  \textbf{Money or gift cards}
  \end{quote}
\item
  \begin{quote}
  \textbf{Authorship on the paper}
  \end{quote}
\item
  \begin{quote}
  \textbf{Course credit}
  \end{quote}
\item
  \begin{quote}
  \textbf{Other compensation}
  \end{quote}
\item
  \begin{quote}
  \textbf{Volunteer / explicit no compensation}
  \end{quote}
\item
  \begin{quote}
  \textbf{No information}
  \end{quote}
\item
  \begin{quote}
  \textbf{Unsure}
  \end{quote}
\end{itemize}

If they are authors on the paper, put authorship

Example: if they said students were fluent in English and Hindi for a
task involving sentiment analysis of English/Hindi tweets, this is a
claim of expertise.

Example: ``We develop a mechanism to help three volunteers analyze each
collected user manually'' -\/- put other, if that is all they say

Example: If it just says ``we annotated\ldots{}'' then assume it is only
the paper's authors unless otherwise stated.

\subsubsection{Training for human annotators}

Did the annotators receive interactive training for this specific
annotation task / research project? Training involves some kind of
interactive feedback. Simply being given formal instructions or
guidelines is not training. Prior professional expertise is not
training. This is not about the qualifications of the annotator, but the
training for this specific project.

Ex: (link removed) is some training data b/c ``one drill session''

Options include:

\begin{itemize}
\item
  \begin{quote}
  Some training detailed
  \end{quote}
\item
  \begin{quote}
  No information in the paper
  \end{quote}
\item
  \begin{quote}
  Unsure
  \end{quote}
\end{itemize}

Example: It is not considered training if there was prescreening, unless
they were told what they got right and wrong or other debriefing. Not
training if they just gave people with high accuracy more work.

Example: This paper had a minimum acceptable statement for some training
information, with only these lines: ``The labeling was done by four
volunteers, who were \emph{carefully instructed} on the definitions in
Section 3. The volunteers agree on more than 90\% of the labels, and any
labeling differences in the remaining accounts are \emph{resolved by
consensus}.''

\subsubsection{Formal instructions/guidelines:} What documents were the
annotators given to help them? This document you are in right now is an
example of formal instructions with definitions and examples.

\begin{itemize}
\item
  \begin{quote}
  No instructions beyond question text
  \end{quote}
\item
  \begin{quote}
  Instructions include formal definition or examples
  \end{quote}
\item
  \begin{quote}
  No information in paper (or not enough to decide)
  \end{quote}
\item
  \begin{quote}
  Unsure
  \end{quote}
\end{itemize}

Example of a paper showing examples:

\begin{quote}
we asked crowdsourcing workers to assign the ``relevant'' label if the
tweet conveys/reports information useful for crisis response such as a
report of injured or dead people, some kind of infrastructure damage,
urgent needs of affected people, donations requests or offers, otherwise
assign the ``non-relevant'' label
\end{quote}

Ex (link removed): no instructions beyond question text

\subsubsection{Multiple annotator overlap:} Did the annotators label at least
some of the same items?

\begin{itemize}
\item
  \begin{quote}
  Yes, for all items
  \end{quote}
\item
  \begin{quote}
  Yes, for some items
  \end{quote}
\item
  \begin{quote}
  No
  \end{quote}
\item
  \begin{quote}
  Unsure
  \end{quote}
\item
  \begin{quote}
  No information
  \end{quote}
\item
  \begin{quote}
  N/A
  \end{quote}
\end{itemize}

If it says there was overlap but not info to say all or some, put unsure
and flag for discussion.

This is usually no for survey/self reported, because it often doesn't
make sense. But could be, e.g. double checking if motion capture
participation participants do the right motions.

\subsubsection{Synthesis of annotator overlap (e.g. majority voting)}

Put N/A if there was no overlap. There might be multiple means stated,
pick the primary synthesis method.

\begin{itemize}
\item
  \begin{quote}
  Qualitative / discussion
  \end{quote}

  \begin{itemize}
  \item
    \begin{quote}
    Process should involve some interaction between labelers
    \end{quote}
  \end{itemize}
\item
  \begin{quote}
  Quantitative / no discussion
  \end{quote}

  \begin{itemize}
  \item
    \begin{quote}
    Includes majority vote, Bayesian, recruit a tiebreaker
    \end{quote}
  \end{itemize}
\item
  \begin{quote}
  Other
  \end{quote}
\item
  \begin{quote}
  No information
  \end{quote}
\item
  \begin{quote}
  Unsure
  \end{quote}
\item
  \begin{quote}
  N/A -\/- if there was no multiple overlap
  \end{quote}
\end{itemize}

Use the default synthesis mechanism. For example: ``Labelers met in
person to discuss cases of disagreement. Where consensus could not be
reached, the project lead made the final decision.'' Put qualitative /
discussion.

Example of bare minimum for Qualitative / discussion: ``resolve{[}d{]}
disagreements by discussing a consensus annotation.''

\subsubsection{Reported inter-annotator agreement}

Leave blank if there was no overlap. Is a metric of inter-annotator
agreement or intercoder reliability reported? It may be called
Krippendorf's alpha, Cohen's kappa, F1 score, or other things. This can
also be in stating that in reconciliation, only items were kept if all
annotators agreed (ex:(link removed)). Borderline in the other
direction: (link removed) -\/- they discussed, but didn't report
pre-discussion agreement.

\begin{itemize}
\item
  \begin{quote}
  Yes
  \end{quote}
\item
  \begin{quote}
  No
  \end{quote}
\item
  \begin{quote}
  Unsure
  \end{quote}
\item
  \begin{quote}
  N/A -\/- if there was no multiple overlap
  \end{quote}
\end{itemize}

\subsubsection{Total number of total human annotators who annotated items}

How many total individuals were involved in evaluating/labeling items?
Think of this like ``number of annotators on the team.'' If it says
there were initially 2 annotators, but a third joined halfway through,
just put 3. This needs to be a single machine-readable number.

\emph{If you cannot answer the question because you don't know how many
annotators (or it just says how many annotators per item), put a - and
answer in the next question.}

If they just say ``we annotated'', don't count the number of authors.
Put no info.

If they didn't use human annotation, put N/A.

For surveys, total would be the number of respondents surveyed.

If all the info you have is an implicit ``we labeled'' (or other info
that it was the authors), this is not sufficient information to answer
the question.

\subsubsection{Median number of human annotators per item}

For papers that had multiple annotator overlap, when multiple annotators
were involved in annotating the same item, how many annotated each item?
Note that some papers will involve a smaller portion of items annotated
by multiple annotators, and a larger portion annotated by just one.
Exclude the cases where only one annotator annotated and determine the
median number of annotators involved.

\emph{If there is no info on this (or it just says total number), put a
-. }

If they just say ``we annotated'', don't count the number of authors.
Put no info.

Example: 75\% of items labeled by 1 person, 20\% labeled by 3 people,
5\% labeled by 10 people. Put 3. This needs to be a single
machine-readable number.

For surveys, median would be 1.

Example with bare minimum (physician's assessment): (link removed)

\subsubsection{Link to dataset available:} Is there a link in the paper to the
human-labeled dataset they used?

\begin{itemize}
\item
  \begin{quote}
  Yes
  \end{quote}
\item
  \begin{quote}
  Yes, but link is broken
  \end{quote}
\item
  \begin{quote}
  No
  \end{quote}
\item
  \begin{quote}
  Unsure
  \end{quote}
\item
  \begin{quote}
  N/A if not using a human-labeled dataset
  \end{quote}
\end{itemize}

Only follow the link in the paper to see if it is broken, you don't need
to verify it is actually a dataset.

}}
\end{document}